%% file: iclr2027_conference.tex
\documentclass{article} 
\usepackage{iclr2027_conference,times}

\input{math_commands.tex}

\usepackage{hyperref}
\usepackage{url}

\usepackage{subcaption}
\usepackage{multirow}
\usepackage{graphicx}
\usepackage{mathtools}
\usepackage{wrapfig}
\usepackage{siunitx}
\usepackage{algorithm, algpseudocode}
\usepackage{booktabs}

\title{Encoding Memory in World Models with Test-Time Model-Based Reinforcement Learning}

\author{Maxime Burchi, Fayaz Ali Dharejo, Radu Timofte \\
Computer Vision Lab, CAIDAS \& IFI, University of Würzburg, Germany \\
\texttt{maxime.burchi@uni-wuerzburg.de}
}

\iclrpreprint

\begin{document}

\maketitle

\input{sec/0_abstract.tex}
\input{sec/1_introduction.tex}
\input{sec/2_background.tex}
\input{sec/3_method.tex}
\input{sec/4_experiments.tex}
\input{sec/6_conclusion.tex}

\bibliography{iclr2027_conference}
\bibliographystyle{iclr2027_conference}

\newpage
\appendix
\section{Appendix}
\input{sec/7_appendix.tex}

\end{document}

%% file: math_commands.tex
\usepackage{amsmath,amsfonts,bm}

\def\eqref#1{equation~\ref{#1}}

\def\1{\bm{1}}

\DeclareMathAlphabet{\mathsfit}{\encodingdefault}{\sfdefault}{m}{sl}
\SetMathAlphabet{\mathsfit}{bold}{\encodingdefault}{\sfdefault}{bx}{n}



%% file: sec/0_abstract.tex
\begin{abstract}
Model-based reinforcement learning (MBRL) approaches have demonstrated strong performance across diverse domains in recent years, learning accurate world models from agent trajectories. However, despite this progress, current approaches still struggle to surpass human performance on tasks that require long-term memory and planning. In this work, we address the problem of long-term memory in MBRL by moving beyond the commonly studied paradigm of architecture design. We introduce Test-Time Model-Based Reinforcement Learning (TT-MBRL), using gradient descent at test time to encode the memory of episode observations into world model parameters. This enables the agent to learn a distinct world model for each episode, improving memory capabilities compared to traditional approaches that solely rely on contextual memory. We first establish a proof of concept on a simplified two-dimensional variant of the Memory Maze environment, which we refer to as Memory Maze 2D, showing consistent improvements across different world model architectures. We then demonstrate the effectiveness of TT-MBRL in the 3D domain, achieving new state-of-the-art results on the Memory Maze benchmark. Finally, we show that TT-MBRL maintains strong performance on non-memory benchmarks such as DeepMind Control and Atari 100k. We will release our code at~\url{https://github.com/burchim/TT-MBRL}.
\end{abstract}

%% file: sec/1_introduction.tex
\section{Introduction}

\begin{wrapfigure}{r}{0.37\textwidth}
    \centering
    \vspace{-2ex}
    \includegraphics[width=\linewidth]{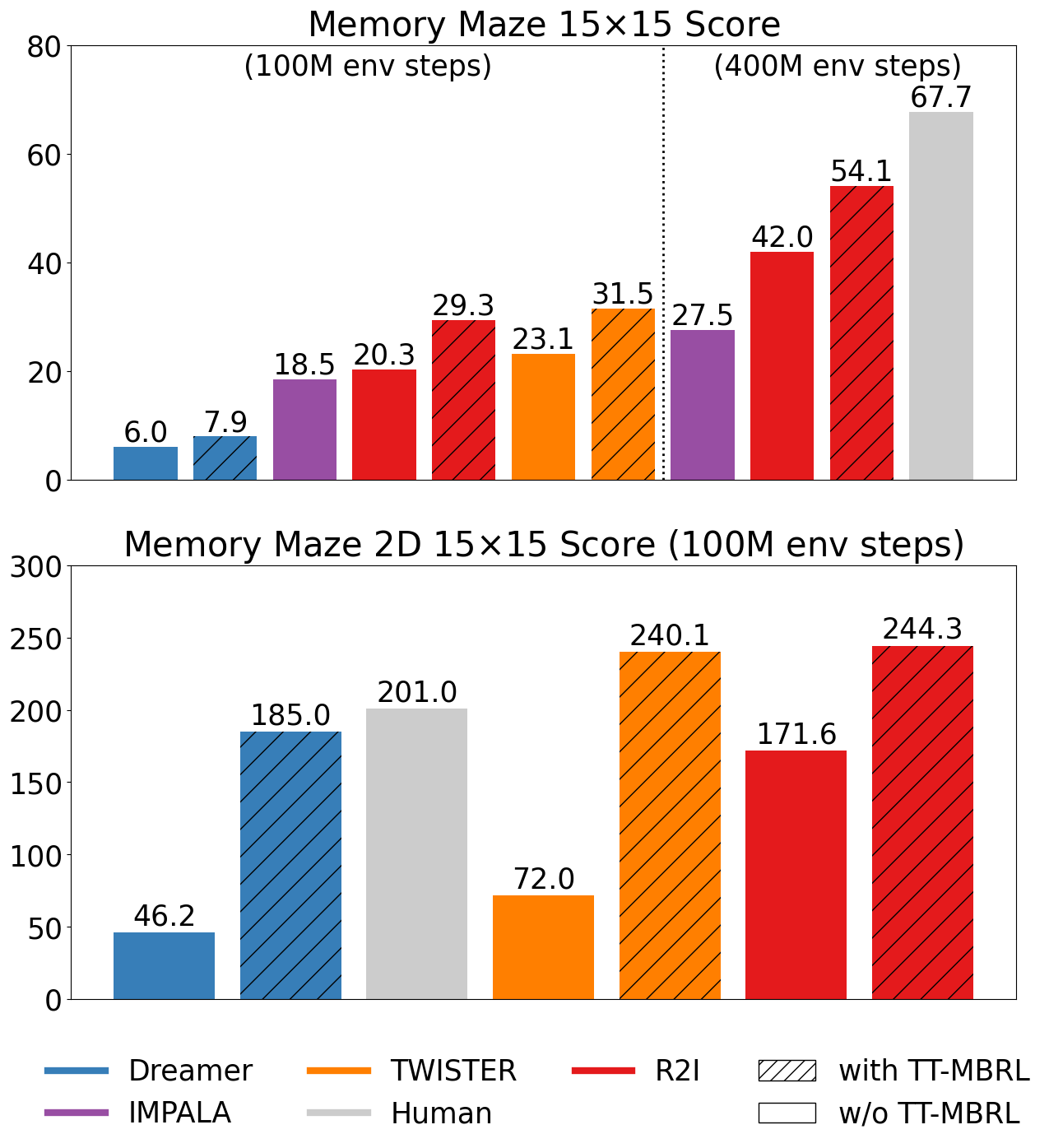}
    \vspace{-3ex}
    \caption{Performance improvements brought by TT-MBRL on Memory Maze 2D and Memory Maze.}
    \vspace{-4ex}
    \label{figure:bar_results}
\end{wrapfigure}

Model-based reinforcement learning has gained increasing research attention in recent years~\citep{kaiser2019model, schrittwieser2020mastering, hafner2019dream, hafner2020mastering, hafner2023mastering, hafner2025training, hansen2022temporal, hansen2024td, hansen2026learning, micheli2024efficient, alonso2024diffusion}, driven in part by advances in hardware computing capabilities. These improvements have enabled the training of world models from high-dimensional observations using deep neural networks as function approximators~\citep{hafner2019learning}. World models~\citep{sutton1991dyna, ha2018recurrent} capture an agent’s experience into a predictive function that can be used in place of the real environment to learn complex behaviors. By leveraging a model of the environment, agents can simulate multiple possible future trajectories in parallel, leading to improved generalization, greater sample efficiency, and more effective decision-making via planning. 

\begin{wrapfigure}{r}{0.3\textwidth}
    \centering
    \vspace{-3ex}
    \includegraphics[width=\linewidth]{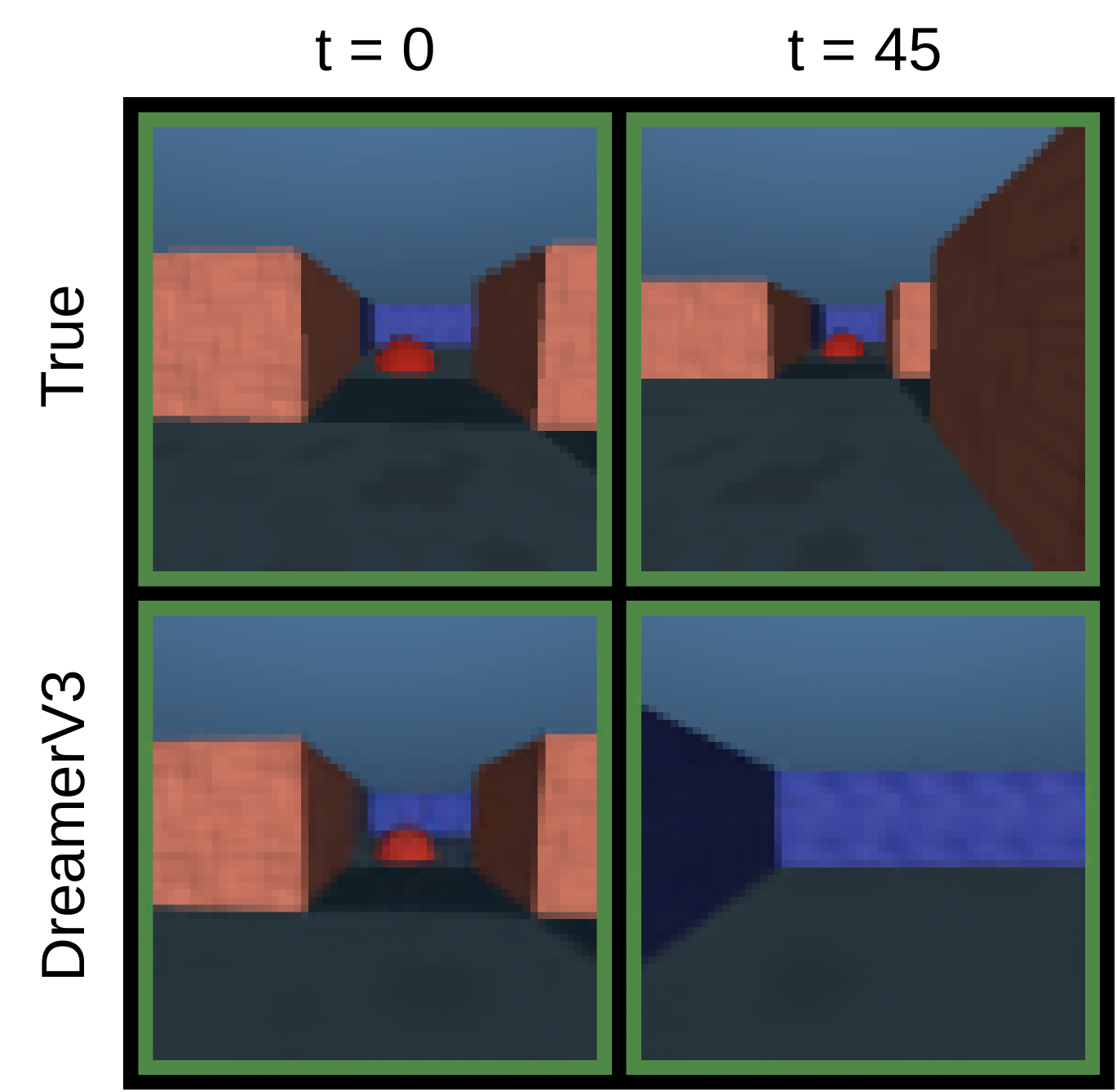}
    \caption{DreamerV3 failure case where the world model predicts an incoherent future after returning to the original position in Memory Maze.}
    \vspace{-2.5ex}
    \label{figure:dreamerv3_fail}
\end{wrapfigure}

Among these approaches, the Dreamer algorithm~\citep{hafner2019dream, hafner2020mastering, hafner2023mastering} has demonstrated strong performance across diverse environments by learning a world model capable of generating accurate trajectories in latent space. However, despite making realistic predictions for fully observable settings, Dreamer struggles to maintain temporal coherence in partially observable environments like Memory Maze (Figure~\ref{figure:dreamerv3_fail}). World model prediction error accumulates over time, leading to long-horizon trajectories that become increasingly inconsistent with past observations.

To remedy this problem, recent approaches have proposed architectural improvements to better handle past context information and enhance the memory capabilities of world models. \citet{pasukonis2023evaluating} proposed to augment Dreamer with Truncated Backpropagation Through Time (TBTT), conditioning training sequences on previous context representations. TransDreamer~\citep{chen2021transdreamer} proposed a Transformer-based world model to bypass the limitations of recurrent neural networks and process long context with self-attention~\citep{vaswani2017attention}. Building on this direction, TWISTER~\citep{burchi2025learning} further improved Transformer-based world models by incorporating a contrastive predictive coding objective to enhance temporal representation learning. Following the success of State Space Models (SSMs) in capturing dependencies across long sequences~\citep{gu2022efficiently, nguyen2022s4nd, smith2023simplified}, several works have also explored SSM-based world models. Recall to Imagine (R2I)~\citep{samsami2024mastering} improved training efficiency and long-term memory using a modification of the S4 model~\citep{gu2022efficiently}, outperforming previous approaches on the Memory Maze benchmark~\citep{pasukonis2023evaluating}. Concurrently, \citet{deng2023facing} proposed S4WM, further demonstrating that SSM-based architectures outperform Transformer-based world models in terms of long-term memory.

However, despite this progress, existing approaches still encounter difficulties surpassing human performance on tasks requiring long-term memory and planning~\citep{samsami2024mastering}. Most prior works focus on architecture design, assuming that improved world model architectures can effectively exploit past context to capture long-term dependencies. These approaches rely on learning a single world model that generalizes across episodes, requiring it to extract common patterns and adapt to the unique structure of each new episode through contextual representations. While shared patterns can be learned effectively, modeling episode-specific information through context alone remains challenging, particularly in partially observable settings like Memory Maze. In contrast, biological systems are believed to encode long-term memory through persistent modifications of synaptic strengths, a process often associated with Long-Term Potentiation (LTP) in neural circuits~\citep{bliss1993synaptic}. Motivated by this perspective, and by the observation that Test-Time Training (TTT) can improve predictive models when training and test data come from different distributions~\citep{sun2020test}, we propose a learning strategy that encodes episode-specific memory directly into the model parameters via gradient-based updates at test time. This approach enables the agent to capture episode-specific information by adapting its world model online, rather than relying solely on contextual memory.

In this work, we address the challenge of long-term memory in MBRL by moving beyond the commonly studied paradigm of architecture design. Inspired by test-time training methods that adapt model parameters at inference time~\citep{sun2020test, gandelsman2022test, wang2025test}, we introduce Test-Time Model-Based Reinforcement Learning (TT-MBRL), using gradient descent at test time to encode the memory of episode-specific information into world model parameters. This enables the agent to learn a distinct world model for each episode, improving memory capabilities compared to traditional approaches that solely rely on contextual memory. We first establish a proof of concept on a simplified two-dimensional variant of the Memory Maze environment, which we refer to as Memory Maze 2D, showing consistent improvements across different world model architectures. We then demonstrate the effectiveness of TT-MBRL in the 3D domain, achieving new state-of-the-art results on the Memory Maze benchmark~\citep{pasukonis2023evaluating}. We find that the effectiveness of TT-MBRL scales proportionally to the number of environment steps collected at test-time, increasing the amount of rewards obtained per step by 53.7\% for R2I after 4000 environment steps on Memory Maze 15\texttimes 15 (Figure~\ref{figure:kl_hist_impr_3d_3}). Finally, we show that TT-MBRL maintains strong performance on non-memory benchmarks such as DeepMind Control and Atari 100k. Figure~\ref{figure:bar_results} summarizes the results obtained on memory benchmarks.

%% file: sec/2_background.tex
\section{Background}

We evaluate the impact of TT-MBRL within the Dreamer family of world model architectures. The Dreamer algorithm was originally introduced in~\citet{hafner2019dream} and later refined in subsequent works~\citep{hafner2020mastering, hafner2023mastering} to master diverse environments with fixed hyperparameters. It has since become a prominent baseline in MBRL, demonstrating strong performance across a wide range of tasks from high-dimensional visual observations. The agent consists of three main components: The world model that learns to simulate the environment to generate imaginary trajectories, the critic network that estimates state values, and the actor network that learns to select optimal actions.

\subsection{World Model Learning}

Dreamer learns a world model of the environment in latent space using a sequential autoencoder architecture~\citep{hinton2006reducing, kingma2014auto, chung2015recurrent}. Specifically, the world model encodes observations $o_{t}$ into compact stochastic representations $z_{t}$ learned via a reconstruction objective, while a Recurrent State-Space Model (RSSM)~\citep{hafner2019learning} predicts future states $z_{t+1}$ given actions $a_{t}$. In addition, the world model includes multi-layer perceptron (MLP) heads for predicting rewards $r_{t}$ and episode continuation $c_{t}$ signals from model states $s_{t}=\{h_{t}, z_{t}\}$. The trainable subcomponents of the world model are the following:
\begin{equation}
\setlength{\tabcolsep}{3pt}
\begin{tabular}{llll}
    \multirow{3}{*}{RSSM $\begin{dcases*} \\ \\ \end{dcases*}$} & Sequential Network: &  & $h_{t} = f_{\phi}(h_{t-1}, z_{t-1}, a_{t-1})$    \\
    & Encoder Network: & & $z_{t} \sim q_{\phi}(z_{t}\ |\ h_{t}, o_{t})$   \\
    & Dynamics Predictor: & & $\hat{z}_{t} \sim p_{\phi}(\hat{z}_{t}\ |\ h_{t})$  \\
    & Reward Predictor: & & $\hat{r}_{t} \sim p_{\phi}(\hat{r}_{t}\ |\ s_{t})$ \\
    & Continue Predictor: & & $\hat{c}_{t} \sim p_{\phi}(\hat{c}_{t}\ |\ s_{t})$ \\
    & Decoder Network: & & $\hat{o}_{t} \sim p_{\phi}(\hat{o}_{t}\ |\ s_{t})$\\
\end{tabular}
\end{equation}
While the original Dreamer algorithm uses a Gated Recurrent Unit (GRU)~\citep{cho2014learning} for the sequential network, other variants have been proposed using Transformer-based world models with self-attention~\citep{vaswani2017attention} and Structured State Space Sequence models (S4)~\citep{gu2022efficiently}. These variants allow computing deterministic states $h_{1:T}$ in parallel using a non-recurrent encoder network $z_{t} \sim q_{\phi}(z_{t}\ |\ o_{t})$ and a parallelizable sequential network $h_{1:T}, x_{1:T}=f_{\phi}(z_{0:T-1}, a_{0:T-1}, x_{0})$, where $x_{t}$ are SSM hidden states.
In this paper, we experiment with three recently proposed world model designs based on the Dreamer algorithm: DreamerV3~\citep{hafner2023mastering}, which uses an RNN-based world model; TWISTER~\citep{burchi2025learning}, which uses a Transformer-based world model with self-attention; and R2I~\citep{samsami2024mastering}, which introduced the Structured State-Space Model (S3M), a modification of S4 to handle temporal dependencies.

The world model is trained on trajectories sampled from a replay buffer. Trajectories are collected during the exploration phase, where the current policy interacts with the environment. Given an input batch containing $B$ sequences of $T$ image observations $o_{1:T}$, actions $a_{1:T}$ , rewards $r_{1:T}$, and episode continuation flags $c_{1:T}$, the world model parameters ($\phi$) are optimized to minimize the following loss function:
\begin{equation}
L(\phi) = \frac{1}{BT}\sum_{b=1}^{B}\sum_{t=1}^{T}\Bigr[\beta_{\text{pred}}L_{\text{pred}}(\phi) + \beta_{\text{dyn}}L_{\text{dyn}}(\phi) + \beta_{\text{rep}}L_{\text{rep}}(\phi)\Bigl]
\end{equation}
The prediction loss $L_{\text{pred}}$ learns hidden representations with a reconstruction loss and predicts environment quantities such as rewards and continuation flags:
\begin{equation}
L_{\text{pred}}(\phi) = \underbracket[0.1ex]{-\log p_{\phi}(o_{t}\ |\ s_{t})}_{\text{reconstruction loss}} \underbracket[0.1ex]{-\log p_{\phi}(r_{t}\ |\ s_{t})}_{\text{reward predictor loss}}\ \underbracket[0.1ex]{-\log p_{\phi}(c_{t}\ |\ s_{t})}_{\text{continue predictor loss}}
\end{equation}
The dynamics loss $L_{\text{dyn}}$ trains the dynamics predictor to predict the next state representation by minimizing the KL divergence between the predictor $p_{\phi}(\hat{z}_{t}\ |\ h_{t})$ and the next stochastic representation $q_{\phi}(z_{t}\ |\ h_{t}, o_{t})$. The representation loss trains the representations to become more predictable if the dynamics predictor cannot predict their distribution. Both loss terms use the stop gradient operator $sg(\cdot)$ to prevent the gradients of targets from being backpropagated and are scaled by the corresponding loss weights $\beta_{\text{dyn}}$ and $\beta_{\text{rep}}$, respectively:
\begin{equation}
\setlength{\tabcolsep}{2pt}
\renewcommand{\arraystretch}{1.5}
\begin{tabular}{rll}
$L_{\text{dyn}}(\phi)=$ & $\max\bigl(1, \text{KL}\bigl[\ sg( q_{\phi}(z_{t}\ |\ h_{t}, o_{t}))\ ||\hspace{3.6ex}p_{\phi}(\hat{z}_{t}\ |\ h_{t})\hspace{0.87ex}\bigr]\bigr)$ \\
$L_{\text{rep}}(\phi)=$ & $\max\bigl(1, \text{KL}\bigl[\hspace{3.6ex}q_{\phi}(z_{t}\ |\ h_{t}, o_{t})\hspace{1.42ex} ||\  sg(  p_{\phi}(\hat{z}_{t}\ |\ h_{t}) )\bigr]\bigr)$
\end{tabular}
\end{equation}

\subsection{Actor-Critic Learning}

The agent’s policy and value functions are learned by imagining trajectories using the world model. Learning takes place entirely in latent space, which allows the agent to process large batch sizes and increase generalization. To do so, the model states of the sampled sequence are flattened along batch and time dimensions to generate $B_{img} = B \times T$ sample trajectories. The world model imagines $H = 15$ steps into the future using the sequential network and dynamics predictor, selecting actions by sampling from the actor network. Analogously to world model predictor networks, actor and critic networks are designed as simple MLPs with parameter vectors ($\theta$) and ($\psi$), respectively. We describe the actor-critic learning process of the agent in more detail in Appendix~\ref{appendix:actor_critic}.

%% file: sec/3_method.tex
\section{Method}
\label{section:method}

\begin{figure}[ht]
    \centering
    \includegraphics[width=\linewidth]{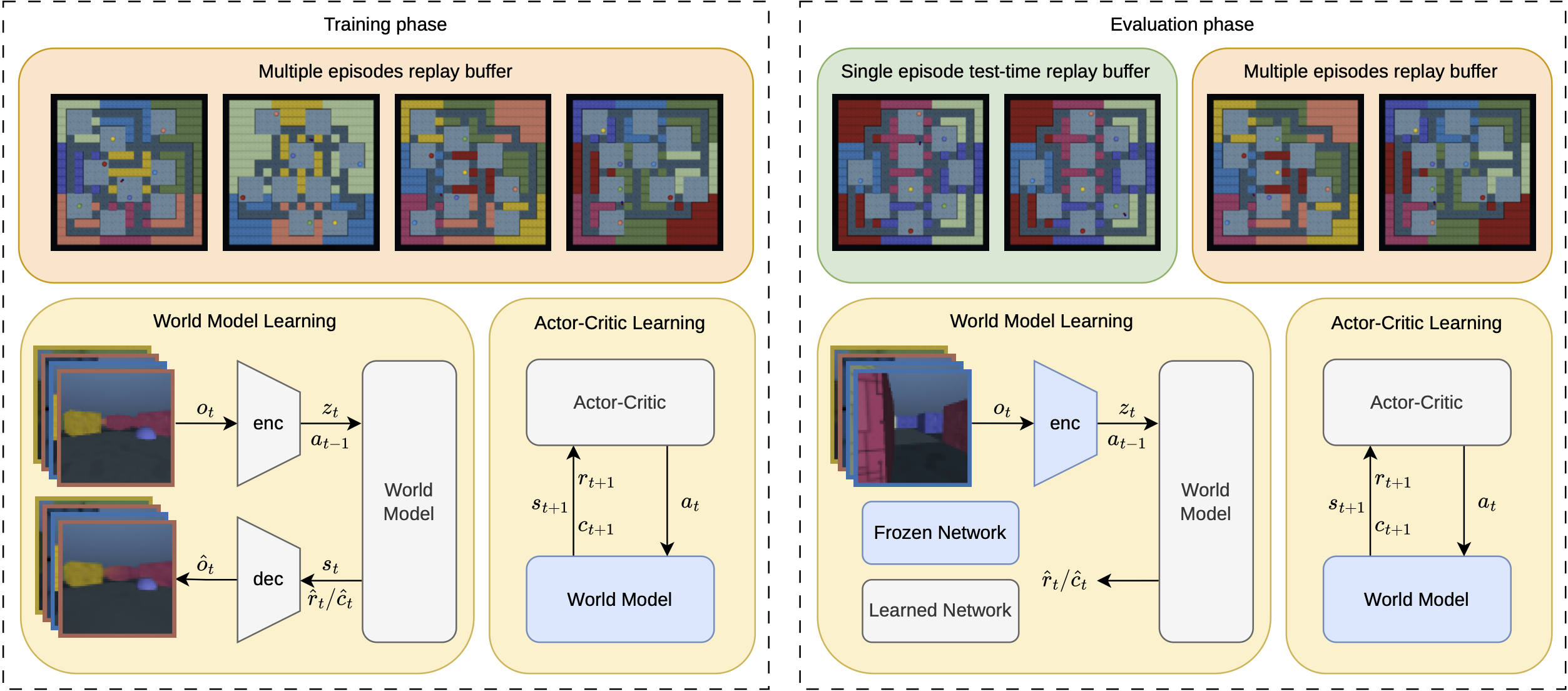}
    \caption{During training, the world model learns to predict the common patterns shared among multiple episode trajectories collected during exploration. The agent can solely rely on contextual memory to identify patterns that are unique to each new episode, which can lead to incoherent future predictions. At test time, the agent can leverage both contextual and neural memory. This allows the world model to make more accurate predictions and learn an improved policy for each new episode.}
    \label{figure:TT-MBRL}
\end{figure}

We introduce TT-MBRL, a novel learning framework for MBRL that takes advantage of test-time observations to learn episode-specific patterns and acquire long-term memory. TT-MBRL learns a unique world model and policy for each evaluation episode. In contrast to training time, where model parameters are shared across all episodes in the replay buffer, this allows the agent to fine-tune its parameters to each individual episode. An overview of the proposed method is provided in Figure~\ref{figure:TT-MBRL}.

\subsection{Test-Time Optimization}

At test time, the agent initializes test-time network parameters ($\phi'$, $\theta'$, $\psi'$) with the set of weights learned during the training phase ($\phi$, $\theta$, $\psi$) at the beginning of each episode. A dedicated test-time replay buffer is also created to store observations collected within the current evaluation episode. Analogous to the training procedure, the agent samples mini-batches of trajectories to update its parameters. We find it beneficial to draw half of the trajectories from the training replay buffer and the other half from the test-time buffer. This strategy acts as a form of regularization, mitigating overfitting to the limited test-time observations, which could otherwise degrade performance. When the number of test-time samples is smaller than the required world model sequence length, mini-batches are concatenated with trajectories from the training buffer. 
We also freeze the encoder and decoder parameters and drop the reconstruction loss at test time to improve TT-MBRL efficiency, preserving the representation space learned during training. Unlike the sequential and dynamics networks, which must adapt to the specific layout and structure of each new episode, the encoder and decoder learn general visual representations that are independent of their temporal ordering. Freezing these parameters significantly reduces evaluation time without impacting performance. We provide an empirical analysis of these design choices in Section~\ref{subsection:analysis}.
For each new policy step, the world model loss optimizes the test-time parameters $\phi'$ as follows:
\begin{equation}
L(\phi') = \frac{1}{BT}\sum_{b=1}^{B}\sum_{t=1}^{T}\Bigr[\beta_{\text{pred}}L_{\text{pred}}(\phi') + \beta_{\text{dyn}}L_{\text{dyn}}(\phi') + \beta_{\text{rep}}L_{\text{rep}}(\phi')\Bigl]
\end{equation}
With the test-time prediction loss $L_{\text{pred}}(\phi')$ omitting the reconstruction objective:
\begin{equation}
L_{\text{pred}}(\phi') = \underbracket[0.1ex]{-\log p_{\phi'}(r_{t}\ |\ s_{t})}_{\text{reward predictor loss}}\ \underbracket[0.1ex]{-\log p_{\phi'}(c_{t}\ |\ s_{t})}_{\text{continue predictor loss}}
\end{equation}
The agent's behavior at test time is optimized using the adapted world model to learn an improved policy. By reducing world model prediction errors, TT-MBRL enables more accurate estimation of imagined returns, which in turn allows the actor and critic to be more effectively optimized.

%% file: sec/4_experiments.tex
\section{Experiments}

In this work, we investigate the potential of TT-MBRL to learn long-term memory via gradient descent. To this end, we evaluate our method on several tasks that require long-term memory and planning. We first establish a proof of concept by experimenting with a simplified two-dimensional variant of the Memory Maze environment, which we refer to as Memory Maze 2D, showing consistent improvements across different world model architectures. We then demonstrate the effectiveness of TT-MBRL in the 3D domain, achieving new state-of-the-art results on the Memory Maze benchmark~\citep{pasukonis2023evaluating}. Finally, we conduct a study to identify effective strategies for performing MBRL at test time and show that TT-MBRL maintains strong performance on non-memory benchmarks such as the DeepMind Control Suite (DMC)~\citep{tassa2018deepmind} and Atari 100k~\citep{kaiser2019model}.
\begin{figure*}[ht]
    \begin{subfigure}{0.33\textwidth}
        \centering
        \includegraphics[width=0.9\linewidth]{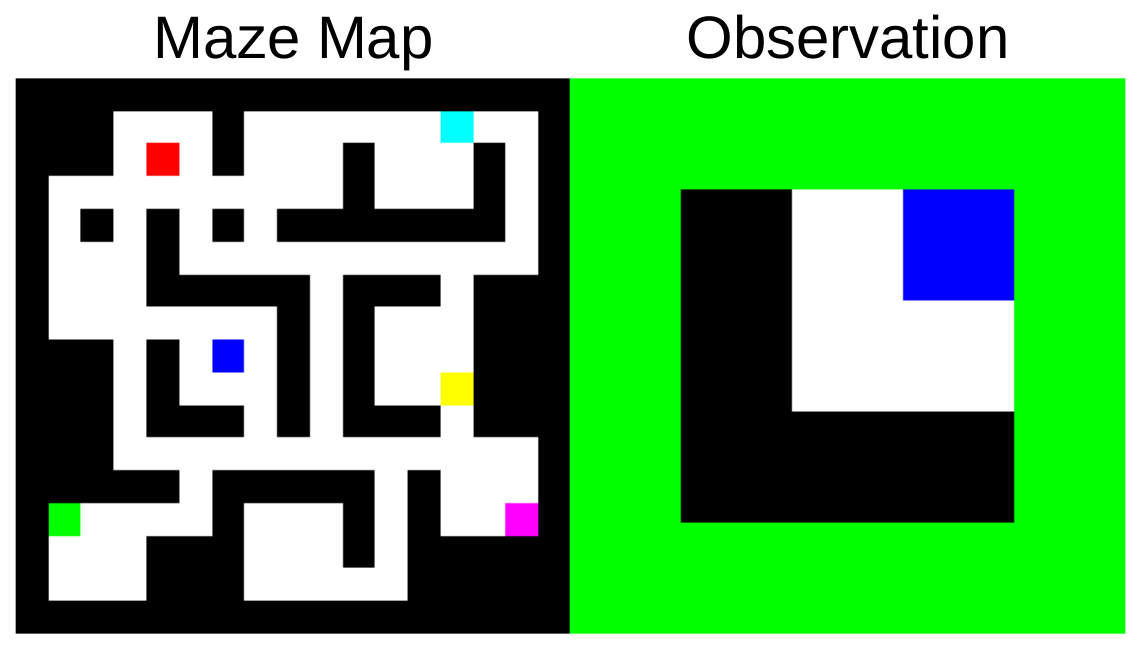}
        \caption{Memory Maze 2D}
        \label{figure:memory_maze_2d}
    \end{subfigure}
    \begin{subfigure}{0.33\textwidth}
        \centering
        \includegraphics[width=0.9\linewidth]{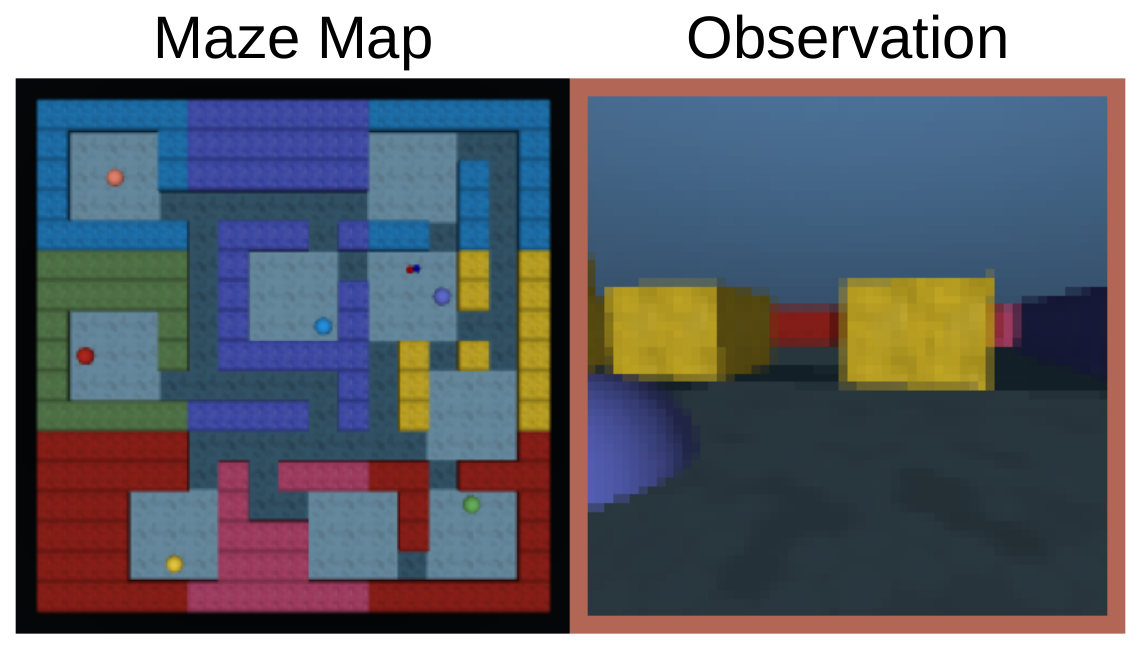}
        \caption{Memory Maze}
    \end{subfigure}
    \begin{subfigure}{0.33\textwidth}
        \centering
        \includegraphics[width=0.9\linewidth]{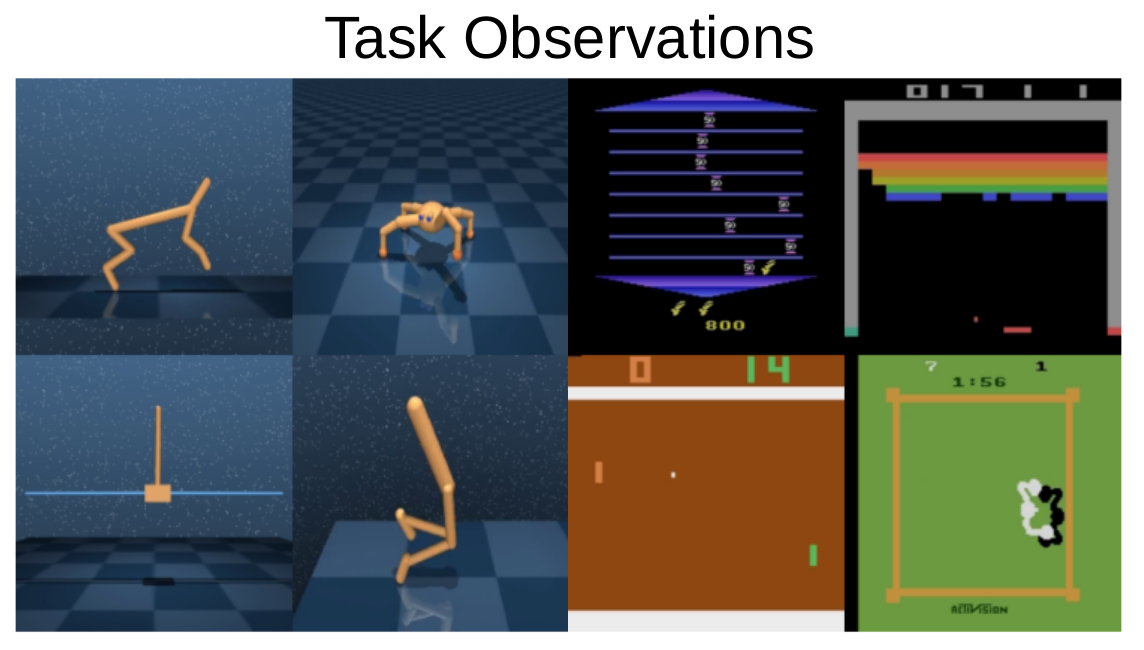}
        \caption{DMC/Atari}
        \label{figure:dmc_atari}
    \end{subfigure}
    \caption{Randomly generated environment map and agent observation for Memory Maze 2D and Memory Maze benchmarks. The agent receives partial observations of the environment and must rely on long-term memory to learn an optimal policy. In contrast, DMC and Atari 100k tasks are mostly fully observable and solvable without extensive memory.}
    \label{figure:memory_maze}
\end{figure*}

\subsection{Memory Maze 2D}
\label{subsection:memory_maze_2d}

\paragraph{Benchmark.}

In order to demonstrate the effectiveness of TT-MBRL, we first design a simplified and efficient two-dimensional variant of the Memory Maze environment in which the agent observes a top-down $3 \times 3$ local context centered on its current position. To enable goal identification, we augment the observation with colored borders that indicate the target color to be reached for obtaining the next reward. The observation is then flattened and normalized, forming a vector of dimension $75=5\times5\times3$. Figure~\ref{figure:memory_maze} illustrates a comparison between our modified setting and the original Memory Maze environment. The agent operates in the maze using a discrete action space consisting of five actions: no-op, move left, move right, move up, and move down. We consider the same four maze sizes as in the original benchmark (9\texttimes9, 11\texttimes11, 13\texttimes13, and 15\texttimes15), along with their corresponding hyperparameters. Further details on the proposed Memory Maze 2D environment are provided in Appendix~\ref{appendix:memory_maze_2d}.

\input{tables/results_maze_2d.tex}
\input{tables/metrics_2d.tex}

\paragraph{Results.}

\begin{figure*}[ht]
    \centering
    \begin{subfigure}{\textwidth}
    \centering
    \includegraphics[width=\linewidth]{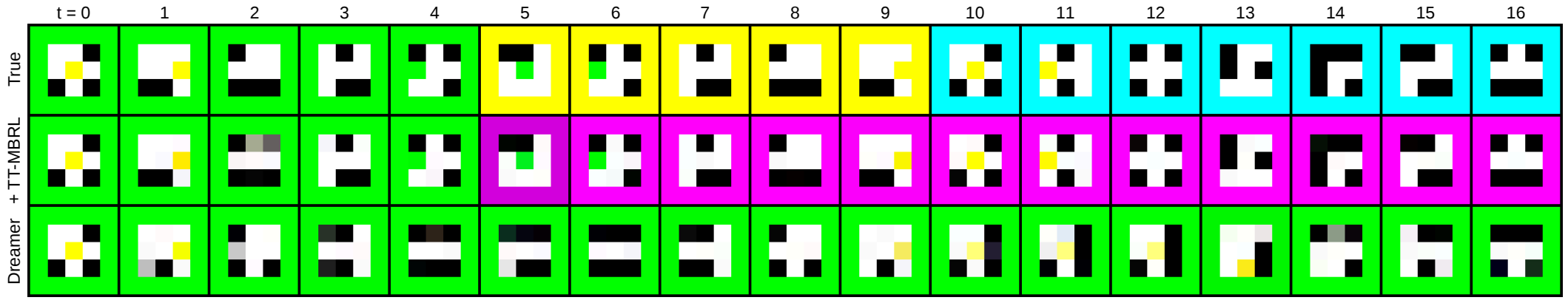}
    \caption{DreamerV3 world model predictions over 16 time steps: Memory Maze 2D 15\texttimes15}
    \label{figure:imagine_2d}
    \end{subfigure}
    \begin{subfigure}{\textwidth}
    \centering
    \includegraphics[width=\linewidth]{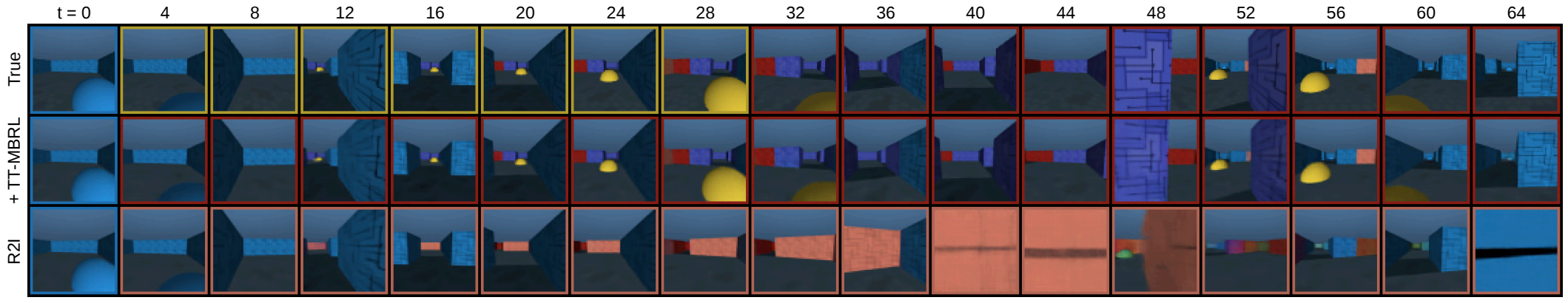}
    \caption{R2I world model predictions over 64 time steps: Memory Maze 15\texttimes15}
    \label{figure:imagine_3d}
    \end{subfigure}
    \caption{Impact of TT-MBRL on world model predictions. We let the agent explore and learn about the environment, then we compare world model predictions to real observations. First row shows unseen real environment observations. Second row shows world model predictions with TT-MBRL. Third row shows world model predictions without TT-MBRL. Without test-time optimization, the model accumulates prediction errors and struggles to make accurate long-term predictions. This makes it difficult to learn an optimal policy for the actor-critic.}
    \label{figure:imagine}
\end{figure*}

To evaluate the effectiveness of TT-MBRL, we consider three recent world model architectures: DreamerV3~\citep{hafner2023mastering}, which employs an RNN-based world model; TWISTER~\citep{burchi2025learning}, which relies on a Transformer-based architecture; and R2I~\citep{samsami2024mastering}, which is built on S4. Table~\ref{table:results_maze_2d} shows the results obtained after training 100M environment steps on each maze size. We also record the performance of a human player to estimate the capabilities of each method compared to human memory abilities. Oracle defines the performance obtained when following the shortest path to the target color. We find that R2I achieves the best results on all maze sizes with a score of 171.6 on Memory Maze 2D 15\texttimes15. TWISTER outperforms DreamerV3 on the 15\texttimes15 maze size due to its Transformer-based world model with self-attention. However, we find that it struggles to learn an optimal policy on smaller maze sizes. We suppose this is due to the nature of the self-attention mechanism that is not biased toward recent observations and requires accurately selecting the needed information at each time step, while DreamerV3 carries past information into a recurrent hidden state. We find that TT-MBRL improves the performance of all models when a near optimal policy is not already achieved. It allows R2I to outperform human performance on all maze sizes, except 9\texttimes9. This is due to the policy occasionally failing on a subset of episodes, despite achieving higher performance on the majority of them. We also evaluate the impact of TT-MBRL on world-model prediction accuracy using a combination of standard video prediction metrics (Table~\ref{table:metrics_2d}).  We measure the world model KL divergence, Fréchet Video Distance (FVD)~\citep{unterthiner2018towards}, Peak Signal-to-Noise Ratio (PSNR), Structural Similarity Index Measure (SSIM)~\citep{wang2004image} and Learned Perceptual Image Patch Similarity (LPIPS)~\citep{zhang2018unreasonable} over 100 unseen trajectories of 64 frames from distinct episodes. We find that TT-MBRL consistently decreases the world model prediction error in latent space. This results in more accurate predictions and better performance in video prediction metrics. Figure~\ref{figure:imagine_2d} shows the impact of TT-MBRL on DreamerV3 world model predictions.

\subsection{Memory Maze}
\label{subsection:memory_maze}

\paragraph{Benchmark.}

The Memory Maze benchmark, introduced by~\citet{pasukonis2023evaluating}, consists of a 3D domain of randomized mazes specifically designed to evaluate long-term memory in agents. In this environment, the agent must locate a target object of a specific color, indicated by the border color in the observation image. Upon reaching and interacting with the correct object, the agent receives a reward of +1, after which a new target object is randomly selected. The benchmark includes four tasks of increasing size and difficulty: Memory Maze 9\texttimes9, 11\texttimes11, 13\texttimes13, and 15\texttimes15. While the smaller maze sizes contain three objects, 15\texttimes15 mazes include six objects and a greater number of rooms, making the task more challenging while remaining solvable for human players.

\input{tables/results_maze.tex}
\input{tables/metrics_3d.tex}

\begin{figure*}[ht]
    \begin{subfigure}{0.33\textwidth}
        \centering
        \includegraphics[width=0.9\linewidth]{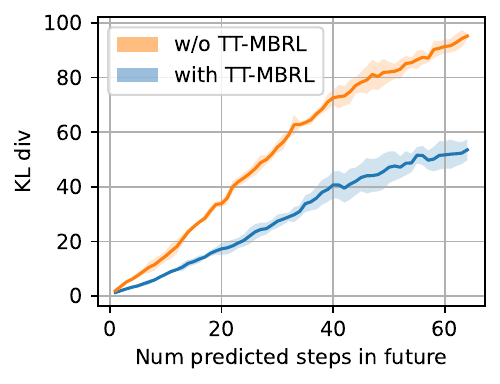}
        \caption{KL div ($\downarrow$) over time steps}
        \label{figure:kl_hist_impr_3d_1}
    \end{subfigure}
    \begin{subfigure}{0.33\textwidth}
        \centering
        \includegraphics[width=0.9\linewidth]{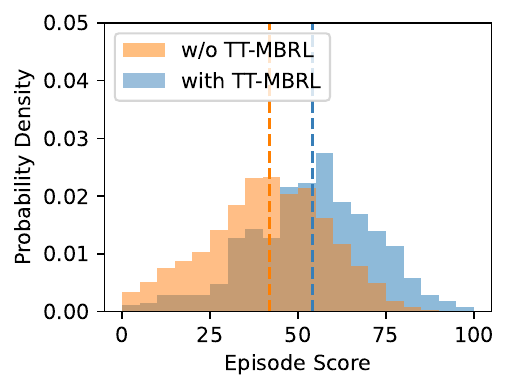}
        \caption{Evaluation histogram}
        \label{figure:kl_hist_impr_3d_2}
    \end{subfigure}
    \begin{subfigure}{0.33\textwidth}
        \centering
        \includegraphics[width=0.9\linewidth]{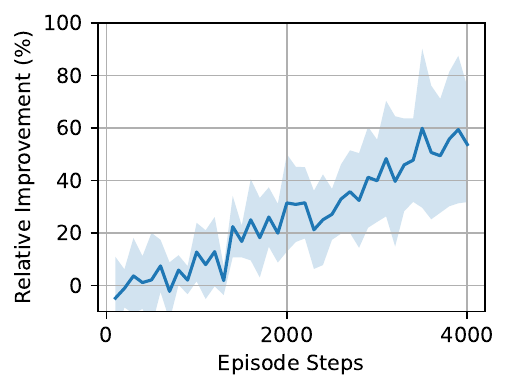}
        \caption{Average obtained reward per step}
        \label{figure:kl_hist_impr_3d_3}
    \end{subfigure}
    \caption{R2I world model KL divergence over 64 predicted steps, evaluation episodes histogram, and relative improvement in obtained rewards brought by TT-MBRL on Memory Maze 15\texttimes 15.}
    \label{figure:kl_hist_impr_3d}
\end{figure*}

\paragraph{Results.}

We train DreamerV3, TWISTER, and R2I world models for 100M environment steps on all Memory Maze sizes. Following~\citet{samsami2024mastering}, we further train R2I up to 400M environment steps and include the results obtained by IMPALA~\citep{espeholt2018impala}, which serves as a strong model-free baseline for comparison. Table~\ref{table:results_maze} compares the results obtained by each method. Despite the increased observation complexity of the 3D domain, where observations of the same position vary with viewing angle, unlike the fixed top-down views of Memory Maze 2D, TT-MBRL successfully identifies structural similarities between observed views to build a more accurate episode-specific world model. TT-MBRL yields consistent improvements across all world model approaches on larger maze sizes, achieving substantial gains: The performance of R2I increases from 60.1 to 63.0 on Memory Maze 13\texttimes 13, and from 42.0 to 54.1 on Memory Maze 15\texttimes 15, establishing new state-of-the-art results and the closest performance to the human baseline (67.7) among all proposed methods. On the smaller maze sizes, where DreamerV3 and R2I already achieve near-optimal performance, the gains are more modest, consistent with the saturation effects observed in Memory Maze 2D. We further assess the quality of world model predictions in Table~\ref{table:metrics_3d}, comparing video prediction metrics. Across all five metrics, TT-MBRL substantially reduces prediction error. Figure~\ref{figure:kl_hist_impr_3d} further illustrates the dynamics of adaptation within a single episode for R2I on Memory Maze 15\texttimes15. Figure~\ref{figure:kl_hist_impr_3d_1} shows the world model KL divergence over 64 predicted steps: Without TT-MBRL, prediction error accumulates rapidly beyond the first 20 steps, while the adapted model maintains substantially lower divergence. Figure~\ref{figure:kl_hist_impr_3d_2} shows the episode return distribution, with TT-MBRL shifting mass toward higher scores and reducing the frequency of low-reward episodes. Figure~\ref{figure:kl_hist_impr_3d_3} shows the per-step relative improvement in obtained rewards brought by TT-MBRL across the episode: The effectiveness of TT-MBRL grows steadily as more observations accumulate in the test-time replay buffer, reaching 53.7\% after 4000 environment steps.

\subsection{TT-MBRL Analysis}
\label{subsection:analysis}

\paragraph{Training Data Regularization.}

We conduct a set of studies to identify effective strategies for performing model-based reinforcement learning at test time. Table~\ref{table:study_test_train} reports the impact of different sampling schemes on mitigating overfitting.
We first consider a naive approach in which the model is updated solely using trajectories collected at test time, without enforcing a minimum number of samples to prefill the test-time replay buffer before learning begins (Test Data Only, No Prefill). Although this strategy improves performance in many episodes, it also degrades performance in others, as the model overfits to limited early observations, resulting in suboptimal policies.
To address this issue, we introduce a delay mechanism that postpones learning until the test-time replay buffer is populated with a sufficient number of samples (1024 Prefill). This reduces overfitting but introduces a trade-off, as it delays the benefits of adaptation.
To overcome both limitations, we propose mixing trajectories collected at test time with samples from the training replay buffer (Test+Train Data, No Prefill), effectively acting as a form of regularization. This approach mitigates early overfitting while enabling immediate adaptation, leading to more stable and consistent performance improvements.
To verify that the gain in performance is due to TT-MBRL rather than additional inference compute, we also perform a compute-matched evaluation. We study the impact of additional compute on performance when only training trajectories are available (Train Data Only). The comparison confirms that the performance gain is due to TT-MBRL rather than additional compute budget.

\input{tables/study_test_train.tex}

\paragraph{Reconstruction Loss and Parameter Freezing.} Table~\ref{table:freezing} shows the impact of freezing or updating encoder/decoder parameters using the reconstruction loss at test time. We find that the two variants achieve comparable performance across different world model architectures. These findings are as expected, given that encoder representations are independent of their temporal ordering. The episode-specific information is instead captured by the sequential and dynamics network parameters. However, unfreezing these parameters comes at a non-trivial computational cost, as gradient back-propagation must be carried through the decoder and encoder, resulting in longer evaluation times. This is especially the case for Memory Maze, where observations are 64\texttimes64 pixels images. We therefore adopt parameter freezing for encoder and decoder networks as our default setting, as it improves evaluation efficiency without sacrificing performance.

\input{tables/study_freezing.tex}

\subsection{Application of TT-MBRL to Non-Memory Domains}

In order to evaluate the capacity of TT-MBRL to maintain strong performance in non-memory domains, domains not designed primarily to test long-term episodic memory, we evaluate its impact on two widely adopted benchmarks that do not require long-term memory or planning: The visual DeepMind Control Suite (DMC)~\citep{tassa2018deepmind}, which comprises 20 continuous control tasks, and the Atari 100k benchmark~\citep{kaiser2019model}, which includes 26 Atari games with discrete actions. Both benchmarks are largely fully observable and can be solved without extensive memory. We report aggregated mean and median scores across all 20 DMC tasks in Table~\ref{table:results_dmc} and human-normalized scores on Atari 100k in Table~\ref{table:results_atari}. Overall, we find that TT-MBRL does not degrade the performance obtained by DreamerV3. Moreover, we find that TT-MBRL consistently improves performance on \textit{Reacher Hard} and \textit{Finger Turn Hard}, where the agent occasionally fails to accurately reach the target sphere. TT-MBRL mitigates this issue by enabling the agent to better detect and correct such errors. Detailed results for individual tasks are provided in Appendix~\ref{appendix:histograms}.
\input{tables/results_dmc_atari.tex}

%% file: tables/results_maze_2d.tex
\begin{table}[ht]
    \caption{Memory Maze 2D benchmark comparison for all maze sizes (100M env steps). We show the impact of TT-MBRL on the performance for each method. The reported scores are averaged over three runs with 100 evaluation episodes. We also report Human mean scores over 10 episodes.}
    \centering
    \setlength{\tabcolsep}{3.25pt}
    \hfill \break
    \begin{tabular}
    {lcccc}
    \toprule
     & \multirow{2}{*}{\begin{tabular}[c]{c}\textbf{Mem Maze 2D} \\ \textbf{9\texttimes9}\end{tabular}} & \multirow{2}{*}{\begin{tabular}[c]{c}\textbf{Mem Maze 2D} \\ \textbf{11\texttimes11}\end{tabular}} & \multirow{2}{*}{\begin{tabular}[c]{c}\textbf{Mem Maze 2D} \\ \textbf{13\texttimes13}\end{tabular}} & \multirow{2}{*}{\begin{tabular}[c]{c}\textbf{Mem Maze 2D} \\ \textbf{15\texttimes15}\end{tabular}} \\
    \textbf{Method} &  &  &  &  \\
    \midrule
    Human & $126.2$ & $172.0$ & $195.0$ & $201.0$ \\ 
    Oracle & $137.2$ & $215.4$ & $276.6$ & $311.4$ \\
    \midrule
    DreamerV3 (TBTT) & $121.2 \pm 3.0\hphantom{0}$ & $185.9 \pm 11.3$ & $108.0 \pm 8.8\hphantom{0}$ & $\hphantom{0}46.2 \pm 7.6\hphantom{0}$ \\
    ~~+ TT-MBRL (Ours) & $\mathbf{122.1 \pm 1.1\hphantom{0}}$ & $\mathbf{189.4 \pm 6.8\hphantom{0}}$ & $\mathbf{207.2 \pm 10.9}$ & $\mathbf{185.0 \pm 17.6}$ \\
    \midrule
    TWISTER & $\hphantom{0}84.8 \pm 0.8\hphantom{0}$ & $100.0 \pm 5.9\hphantom{0}$ & $\hphantom{0}95.3 \pm 10.2$ & $\hphantom{0}72.0 \pm 3.2\hphantom{0}$  \\
    ~~+ TT-MBRL (Ours) & $\mathbf{113.7 \pm 1.8\hphantom{0}}$ & $\mathbf{178.7 \pm 4.4\hphantom{0}}$ & $\mathbf{224.2 \pm 6.1\hphantom{0}}$ & $\mathbf{240.1 \pm 17.6}$ \\
    \midrule
    R2I & $\mathbf{123.1 \pm 0.2\hphantom{0}}$ & $\mathbf{200.3 \pm 6.3\hphantom{0}}$ & $222.9 \pm 6.6\hphantom{0}$ & $171.6 \pm 24.3$ \\
    ~~+ TT-MBRL (Ours) & $\mathbf{123.1 \pm 0.2\hphantom{0}}$ & $199.2 \pm 4.2\hphantom{0}$ & $\mathbf{233.4 \pm 8.3\hphantom{0}}$ & $\mathbf{244.3 \pm 7.2\hphantom{0}}$ \\
    \bottomrule
    \end{tabular}
    \label{table:results_maze_2d}
\end{table}

%% file: tables/metrics_2d.tex
\begin{table}[ht]
    \caption{Evaluation of the impact of TT-MBRL on world model predictions on Memory Maze 2D 15\texttimes 15. We aggregate all metrics over 100 unseen trajectories of 64 frames from distinct episodes.}
    \centering
    \hfill \break
    \begin{tabular}{l@{\hspace{0.5cm}}c@{\hspace{1.0cm}}c@{\hspace{1.0cm}}c@{\hspace{1.0cm}}c@{\hspace{1.0cm}}c}
    \toprule
    \textbf{Method} & \textbf{KL div $\downarrow$} & \textbf{FVD $\downarrow$} & \textbf{PSNR $\uparrow$} & \textbf{SSIM $\uparrow$} & \textbf{LPIPS $\downarrow$} \\
    \midrule
    DreamerV3 (TBTT) & $114.6$ & $104.5$ & $10.3$ & $0.552$ & $0.283$ \\
    ~~+ TT-MBRL (Ours) & $\mathbf{\hphantom{0}66.0}$ & $\mathbf{\hphantom{0}55.2}$ & $\mathbf{16.3}$ & $\mathbf{0.642}$ & $\mathbf{0.241}$ \\
    \midrule
    TWISTER & $\hphantom{0}28.7$ & $\hphantom{0}89.1$ & $13.1$ & $0.578$ & $0.292$ \\
    ~~+ TT-MBRL (Ours) & $\mathbf{\hphantom{0}13.2}$ & $\mathbf{\hphantom{0}57.1}$ & $\mathbf{15.7}$ & $\mathbf{0.641}$ & $\mathbf{0.262}$ \\
    \midrule
    R2I & $\hphantom{0}52.0$ & $115.4$ & $10.1$ & $0.541$ & $0.304$ \\
    ~~+ TT-MBRL (Ours) & $\mathbf{\hphantom{0}13.8}$ & $\mathbf{\hphantom{0}45.7}$ & $\mathbf{15.8}$ & $\mathbf{0.675}$ & $\mathbf{0.246}$ \\
    \bottomrule
    \end{tabular}
    \label{table:metrics_2d}
\end{table}

%% file: tables/results_maze.tex
\begin{table}[ht]
    \caption{Memory Maze benchmark comparison (100M env steps) for all maze sizes. We show the impact of TT-MBRL on the performance for each method. The reported scores are averaged over three runs with 100 episodes for evaluation. $\dagger$ denotes the results taken from~\citet{samsami2024mastering}. We also report TT-MBRL results for R2I with a budget of 400M environment steps over six runs.}
    \centering
    \setlength{\tabcolsep}{10pt}
    \hfill \break
    \begin{tabular}{lcccc}
    \toprule
     & \multicolumn{1}{c}{\multirow{2}{*}{\begin{tabular}[c]{@{}c@{}}\textbf{Mem Maze} \\ \textbf{9\texttimes9}\end{tabular}}} & \multicolumn{1}{c}{\multirow{2}{*}{\begin{tabular}[c]{@{}c@{}}\textbf{Mem Maze} \\ \textbf{11\texttimes11}\end{tabular}}} & \multicolumn{1}{c}{\multirow{2}{*}{\begin{tabular}[c]{@{}c@{}}\textbf{Mem Maze} \\ \textbf{13\texttimes13}\end{tabular}}} & \multicolumn{1}{c}{\multirow{2}{*}{\begin{tabular}[c]{@{}c@{}}\textbf{Mem Maze} \\ \textbf{15\texttimes15}\end{tabular}}} \\
    \textbf{Method} &  &  &  &  \\
    \midrule
    Human & $26.4$ & $44.3$ & $55.5$ & $67.7$ \\ 
    Oracle & $34.8$ & $58.0$ & $74.5$ & $87.7$ \\
    IMPALA$^{\dagger}$ & $23.5$ & $28.5$ & $27.0$ & $18.5$ \\
    IMPALA$^{\dagger}$ (400M steps) & $26.5$ & $35.0$ & $36.5$ & $27.5$ \\
    \midrule
    DreamerV3 (TBTT) & $\mathbf{32.7 \pm 0.8}$ & $33.0 \pm 1.4$ & $22.1 \pm 1.8$ & $\hphantom{0}6.0 \pm 0.3$\\
    ~~+ TT-MBRL (Ours) & $32.5 \pm 0.7$ & $\mathbf{37.8 \pm 1.1}$ & $\mathbf{32.5 \pm 3.2}$ & $\mathbf{\hphantom{0}7.9 \pm 1.3}$ \\
    \midrule
    TWISTER & $28.2 \pm 0.6$ & $33.5 \pm 0.9$ & $29.7 \pm 2.4$ & $23.1 \pm 0.5$ \\
    ~~+ TT-MBRL (Ours) & $\mathbf{30.6 \pm 0.8}$ & $\mathbf{42.1 \pm 0.3}$ & $\mathbf{42.3 \pm 1.9}$ & $\mathbf{31.5 \pm 1.8}$ \\
    \midrule
    R2I & $\mathbf{32.5 \pm 0.6}$ & $45.9 \pm 1.6$ & $45.9 \pm 1.3$ & $20.3 \pm 1.1$ \\
    ~~+ TT-MBRL (Ours) & $32.3 \pm 0.5$ & $\mathbf{47.8 \pm 2.6}$ & $\mathbf{53.0 \pm 1.8}$ & $\mathbf{29.3 \pm 4.7}$ \\
    \midrule
    R2I (400M steps) & $33.7 \pm 0.3$ & $52.3 \pm 0.9$ & $60.1 \pm 1.6$ & $42.0 \pm 5.8$ \\
    ~~+ TT-MBRL (Ours)& $\mathbf{33.8 \pm 0.6}$ & $\mathbf{52.4 \pm 1.4}$ & $\mathbf{63.0 \pm 1.5}$ & $\mathbf{54.1 \pm 3.5}$ \\
    \bottomrule
    \end{tabular}
    \label{table:results_maze}
\end{table}

%% file: tables/metrics_3d.tex
\begin{table}[ht]
    \caption{Evaluation of the impact of TT-MBRL on world model predictions on Memory Maze 15\texttimes 15. We aggregate all metrics over 100 unseen trajectories of 64 frames from distinct episodes.}
    \centering
    \setlength{\tabcolsep}{11pt}
    \hfill \break
    \begin{tabular}{lccccc}
    \toprule
    \textbf{Method} & \textbf{KL div $\downarrow$} & \textbf{FVD $\downarrow$} & \textbf{PSNR $\uparrow$} & \textbf{SSIM $\uparrow$} & \textbf{LPIPS $\downarrow$} \\
    \midrule
    DreamerV3 (TBTT) & $\hphantom{0}21.7$ & $\hphantom{0}67.7$ & $25.0$ & $0.771$ & $0.211$ \\
    ~~+ TT-MBRL (Ours) & $\hphantom{0}\mathbf{20.1}$ & $\hphantom{0}\mathbf{61.7}$ & $\mathbf{26.2}$ & $\mathbf{0.789}$ & $\mathbf{0.190}$ \\
    \midrule
    TWISTER & $\hphantom{0}51.8$ & $\hphantom{0}63.8$ & $24.6$ & $0.756$ & $0.230$ \\
    ~~+ TT-MBRL (Ours) & $\mathbf{\hphantom{0}46.3}$ & $\mathbf{\hphantom{0}49.1}$ & $\mathbf{26.2}$ & $\mathbf{0.792}$ & $\mathbf{0.187}$ \\
    \midrule
    R2I & $\hphantom{0}61.3$ & $114.1$ & $21.1$ & $0.687$ & $0.319$ \\
    ~~+ TT-MBRL (Ours) & $\mathbf{\hphantom{0}38.3}$ & $\mathbf{85.2}$ & $\mathbf{24.0}$ & $\mathbf{0.750}$ & $\mathbf{0.252}$ \\
    \midrule
    R2I (400M steps) & $\hphantom{0}54.2$ & $111.1$ & $21.7$ & $0.697$ & $0.290$ \\
    ~~+ TT-MBRL (Ours)& $\mathbf{\hphantom{0}29.3}$ & $\mathbf{\hphantom{0}70.7}$ & $\mathbf{26.0}$ & $\mathbf{0.784}$ & $\mathbf{0.192}$ \\
    \bottomrule
    \end{tabular}
    \label{table:metrics_3d}
\end{table}

%% file: tables/study_test_train.tex
\begin{table}[ht]
    \caption{Impact of replay buffer sampling strategy on TT-MBRL performance on Memory Maze 2D 15\texttimes15. Combining training and test-time trajectories (Test+Train Data, No Prefill) consistently achieves the best performance across all world model architectures.}
    \centering
    \setlength{\tabcolsep}{3.5pt}
    \hfill \break
    \begin{tabular}{lcccccc}
    \toprule
    \textbf{Method} & \textbf{w/o TT-MBRL} &\multirow{2}{*}{\begin{tabular}[c]{c}\textbf{Train Data} \\ \textbf{Only}\end{tabular}} & \multicolumn{2}{c}{\textbf{Test Data Only}} & \multicolumn{2}{c}{\textbf{Test+Train Data}} \\
    \cmidrule(lr){4-5} \cmidrule(lr){6-7}
    &&&\textbf{No Prefill} & \textbf{1024 Prefill} & \textbf{No Prefill} & \textbf{1024 Prefill} \\
    \midrule
    DreamerV3 & $\hphantom{0}46.2$ & $\hphantom{0}47.0$ & $117.6$ & $173.5$ & $\mathbf{185.0}$ & $149.6$ \\
    TWISTER & $\hphantom{0}72.0$ & $\hphantom{0}69.2$ & $167.6$ & $210.1$ & $\mathbf{240.1}$ & $207.1$ \\
    R2I & $171.6$ & $171.7$ & $218.8$ & $240.1$ & $\mathbf{244.3}$ & $237.5$ \\
    \bottomrule
    \end{tabular}
    \label{table:study_test_train}
\end{table}

%% file: tables/study_freezing.tex
\begin{table}[ht]
    \caption{Impact of updating encoder/decoder using the reconstruction loss on performance and evaluation time. \textit{Frozen} denotes our default TT-MBRL setting where encoder and decoder parameters are fixed at test time; \textit{Unfrozen} allows them to be updated. We report the evaluation time per episode.}
    \centering
    \setlength{\tabcolsep}{11pt}
    \begin{tabular}{lccccc}
    \toprule
    \textbf{Method} & \textbf{w/o TT-MBRL} & \multicolumn{2}{c}{\textbf{Episode Score}} 
      & \multicolumn{2}{c}{\textbf{Eval. Time (hours)}} \\
    \cmidrule(lr){3-4} \cmidrule(lr){5-6}
    & & \textbf{Frozen} & \textbf{Unfrozen} & \textbf{Frozen} & \textbf{Unfrozen} \\
    \midrule
    \multicolumn{6}{l}{\textit{Memory Maze 2D 15$\times$15}}\\ 
    DreamerV3 & $\hphantom{0}46.2$ & $\mathbf{185.0}$ & $181.4$ & $\mathbf{0.31}$ & $0.35$ \\
    TWISTER   & $\hphantom{0}72.0$ & $240.1$ & $236.4$ & $\mathbf{0.25}$ & $0.27$ \\
    R2I       & $171.6$ & $244.3$ & $\mathbf{247.5}$ & $\mathbf{0.06}$ & $0.06$ \\
    \midrule
    \multicolumn{6}{l}{\textit{Memory Maze 15$\times$15}}\\ 
    R2I       & $42.0$ & $\mathbf{54.1}$ & $53.8$ & $\mathbf{0.65}$ & $1.04$ \\
    \bottomrule
    \end{tabular}
    \label{table:freezing}
\end{table}

%% file: tables/results_dmc_atari.tex
\begin{table}[h]
\centering
\begin{minipage}{0.49\textwidth}
    \centering
    \caption{DMC results (20 tasks, 1M env steps).}
    \begin{tabular}{lcc}
    \toprule
    \textbf{Method} & \textbf{Mean} & \textbf{Median} \\ 
    \midrule
    DreamerV3 & $776.2$ & $842.2$ \\ 
    ~~+TT-MBRL (Ours) & $\mathbf{788.2}$ & $\mathbf{850.3}$ \\
    \bottomrule
    \end{tabular}
\label{table:results_dmc}
\end{minipage}
\hfill
\begin{minipage}{0.49\textwidth}
    \centering
    \caption{Atari results (26 tasks, 400K env steps).}
    \setlength{\tabcolsep}{2.5pt}
    \begin{tabular}{lccc}
    \toprule
    \textbf{Method} & \textbf{Mean} & \textbf{Median} & \textbf{IQM} \\ 
    \midrule
    DreamerV3 & $109.5\%$ & $44.3\%$ & $0.502$ \\ 
    ~~+TT-MBRL (Ours) & $\mathbf{111.1\%}$ & $\mathbf{46.1\%}$ & $\mathbf{0.514}$ \\
    \bottomrule
    \end{tabular}
\label{table:results_atari}
\end{minipage}
\end{table}

%% file: sec/6_conclusion.tex
\section{Conclusion}

In this work, we addressed the challenge of long-term memory in MBRL. While prior approaches have primarily focused on improving world model architectures to better exploit contextual information, we instead explored an orthogonal direction: encoding memory directly into model parameters. To this end, we introduced TT-MBRL, a framework that leverages gradient-based updates at test time to incorporate episode-specific information into the world model. By enabling the agent to learn a distinct set of parameters for each episode, TT-MBRL moves beyond shared representations and allows rapid adaptation to the current episode, effectively forming a form of neural memory. Through experiments on both a simplified setting and the Memory Maze benchmark, we demonstrated that TT-MBRL consistently improves performance across multiple world model architectures. Notably, our method achieves state-of-the-art results on Memory Maze. We find that the effectiveness of TT-MBRL scales proportionally to the number of environment steps collected at test-time. These results highlight the potential of parameter-based memory as a powerful complement to contextual memory in MBRL systems. More broadly, our findings suggest that test-time learning constitutes a promising direction for developing agents capable of long-term planning.

\subsection{Limitations and Future Work.}

The primary limitation of our approach is the reduced evaluation efficiency resulting from the use of gradient-based optimization at test time. TT-MBRL requires learning a distinct world model and behavior policy for each evaluation episode, leading to increased computational cost and longer evaluation times. While this overhead remains manageable in environments such as Memory Maze, it may become prohibitive in settings that require a large number of environment interactions. As a direction for future work, we plan to extend our approach to operate during the exploration phase. We hypothesize that integrating TT-MBRL into exploration could improve sample efficiency and overall performance by enabling the collection of more informative and diverse trajectories. Consequently, improving the computational efficiency of our approach will be crucial to making it practical for online adaptation during exploration.

%% file: sec/7_appendix.tex
\input{sec/2_related_works.tex}

\subsection{Additional Ablations}

\paragraph{Test-Time Optimization Objective.} In order to clarify that the performance gains achieved by TT-MBRL are the results of its key differences rather than solely the product of applying test-time adaptation, we perform a comparison with PAD and study the impact of loss components. Table~\ref{table:study_objective} shows the impact of test-time objectives on performance. As expected, we find that the KL objective is essential to TT-MBRL success. Removing the KL objective (w/o KL loss) causes the agent to fail because the dynamics predictor cannot adapt to moving world model representations and makes incorrect predictions. On the other hand, the agent appears robust to removing the reward and continue objectives (w/o rwd-cont loss), signaling that it is not essential to TT-MBRL success. However, given the dependency of reward and continue MLP heads on moving representations, we suppose this design choice could have a negative impact on performance in the long run. We also find that removing the actor-critic objective (w/o actor-critic) can decrease performance. The performance of the agent deteriorates along the episode, given that the actor and critic cannot adapt to moving representations. We also experiment with PAD~\citep{hansen2021self}, which was originally proposed in the model-free context. PAD only adapts encoder parameters using a self-supervised objective and does not use training samples for regularization, leaving policy and value networks unchanged at test time. As expected, we find that PAD has a detrimental effect on performance despite relying on test-time training.

\input{tables/study_objective.tex}

\paragraph{TT-MBRL Train/Test Sampling Ratio.} The default TT-MBRL setting uses a 50/50 sampling ratio for train/test trajectories. To confirm that our method is robust to this hyper-parameter choice and does not depend on a narrow setting, we study the impact of the train/test ratio by varying the ratio of training samples among the global batch. Table~\ref{table:study_train_test_ratio} shows the results that we obtained on Memory Maze 2D 15\texttimes15. We find that sampling only training trajectories (100/0) does not improve performance because the world model cannot learn about episode-specific patterns. Using only test trajectories (0/100) decreases performance, confirming the important role of training trajectories in regularization. Overall, we find that a sampling ratio between 25\% and 75\% achieves robust performance, showing the flexibility of this hyperparameter.

\input{tables/study_train_test_ratio.tex}

\paragraph{Test-Time Gradient Update Frequency.} We perform a study on compute–performance tradeoff by varying the number of policy steps per gradient step. Table~\ref{table:study_frequency} shows the impact of test-time gradient update frequency on TT-MBRL efficacy.  We find that increasing the update frequency by a factor of 2 does not lead to a significant performance improvement while increasing evaluation time by the same factor. Decreasing the frequency of TT-MBRL updates slows down adaptation, which results in lower final performance. However, we find that TT-MBRL is quite robust to this hyper-parameter, achieving significant improvement with a larger factor of policy steps per gradient step while also decreasing evaluation time by the same factor. We find these results encouraging for the development of an efficient manner to apply TT-MBRL.

\input{tables/study_frequency.tex}

\subsection{Actor Critic Learning}
\label{appendix:actor_critic}

Analogously to world model predictor networks, actor and critic networks are designed as simple MLPs with parameter vectors ($\theta$) and ($\psi$), respectively:
\begin{equation}
    \text{Actor:} \quad a_{t} \sim \pi_{\theta}(a_{t} | s_{t}), \quad \text{Critic:} \quad v_{\psi}(s_{t}) \approx \mathrm{E}_{p_{\phi}, \pi_{\theta}}[R^{\lambda}_{t}]
\end{equation}
The critic network learns to minimize the symlog-transformed cross-entropy loss~\citep{hafner2023mastering} with discretized $\lambda$-returns obtained from imagined trajectories: 
\begin{equation}
\setlength{\tabcolsep}{10pt}
\begin{tabular}{ll}
$R_{t}^{\lambda} = \hat{r}_{t+1} + \gamma \hat{c}_{t+1} \Bigl((1-\lambda)v_{\psi}(s_{t+1}) + \lambda R_{t+1}^{\lambda}\Bigr)$ & $R_{H+1}^{\lambda} = v_{\psi}(s_{H+1})$ 
\end{tabular}
\end{equation}
The critic also relies on its own predictions for estimating rewards beyond the prediction horizon, which requires adding a regularizing term toward the outputs of its own EMA network. Equation~\ref{equation:critic_loss} defines the critic loss as follows:
\begin{equation}
L_{\text{critic}}(\psi) = \frac{1}{BH}\sum_{b=1}^{B}\sum_{t=1}^{H}\Bigl[\ \underbracket[0.1ex]{-\log p_{\psi}(R_{t}^{\lambda} \ |\ s_{t})}_{\text{discrete returns regression}} \underbracket[0.1ex]{-\log p_{\psi}(v'_{\psi}(s_{t}) \ |\ s_{t})}_{\text{critic EMA regularizer}}\ \Bigr]
\label{equation:critic_loss}
\end{equation}
The actor network is trained to select actions that maximize obtained advantages using Reinforce~\citep{williams1992simple}. It also maximizes the policy entropy to ensure sufficient exploration during both data collection and imagination. The actor loss is defined as follows:
\begin{equation}
L_{\text{actor}}(\theta) = \frac{1}{BH}\sum_{b=1}^{B}\sum_{t=1}^{H}\Bigl[\ \underbracket[0.1ex]{-\ \log \pi_{\theta}(a_{t}\ |\ s_{t})sg(A_{t}^{\lambda})}_{\text{reinforce}} \underbracket[0.1ex]{-\ \eta \mathrm{H}[\pi_{\theta}(a_{t}\ |\ s_{t})]}_{\text{entropy regularizer}}\ \Bigr]
\end{equation}
Where $A_{t}^{\lambda}=\big(\hat{R}_{t}^{\lambda} - v_{\psi}(s_{t})\big) / \max(1, S)$ defines advantages computed using normalized returns. The returns are scaled using exponentially moving average statistics of their $5^{th}$ and $95^{th}$ batch percentiles to ensure stable learning across environments with different return scales:
\begin{equation}
S=\operatorname{EMA}(\operatorname{Per}(R_{t}^{\lambda}, 95) - \operatorname{Per}(R_{t}^{\lambda}, 5), momentum=0.99)
\end{equation}

\newpage

\subsection{Experiments Compute Resources}
\label{section:compute}

We perform Memory Maze 2D and Memory Maze experiments on Nvidia L40 GPUs with 48GB of memory along with 40 CPU workers. During training, we use two GPUs for training R2I on Memory Maze experiments while a single GPU is sufficient for other experiments. For DMC and Atari 100K benchmarks, we used RTX 3090 GPUs with 24GB of memory along with 16 CPU workers. A single GPU per task was sufficient for the two benchmarks. Table~\ref{table:compute} describes the execution time and resources required for each experiment.

\begin{table}[ht]
    \caption{Resources and execution time used for each experiments.}
    \centering
    \setlength{\tabcolsep}{5pt}
    \hfill \break
    \begin{tabular}{lcccccc}
    \toprule
     &  &  & & & \multicolumn{1}{c}{\multirow{2}{*}{\begin{tabular}[c]{@{}c@{}}\textbf{\#Env} \\ \textbf{Steps}\end{tabular}}} & \multicolumn{1}{c}{\multirow{2}{*}{\begin{tabular}[c]{@{}c@{}}\textbf{Training Time} \\ \textbf{(days/task)}\end{tabular}}} \\
    \textbf{Environment} & \textbf{\#Tasks} & \textbf{Method} & \textbf{GPU} & \textbf{\#GPUs} \\
    \midrule
    Memory Maze 2D & 4 & DreamerV3 & L40 & 1 & 100M & 8 \\
    Memory Maze 2D & 4 & TWISTER & L40 & 1 & 100M & 6 \\
    Memory Maze 2D & 4 & R2I & L40 & 1 & 100M & 2.9  \\
    Memory Maze & 4 & DreamerV3 & L40 & 1 & 100M & 13.5  \\
    Memory Maze & 4 & TWISTER & L40 & 1 & 100M & 23\\
    Memory Maze & 4 & R2I & L40 & 2 & 400M & 28\\
    DMC & 20 & DreamerV3 & RTX 3090 & 1 & 1M & 0.8 \\
    Atari 100k & 26 & DreamerV3 & RTX 3090 & 1 & 400K & 0.35 \\
    \bottomrule
    \end{tabular}
    \label{table:compute}
\end{table}

\newpage

\subsection{Memory Maze 2D Environment}
\label{appendix:memory_maze_2d}

\begin{figure*}[ht]
    \centering
    \includegraphics[width=0.95\linewidth]{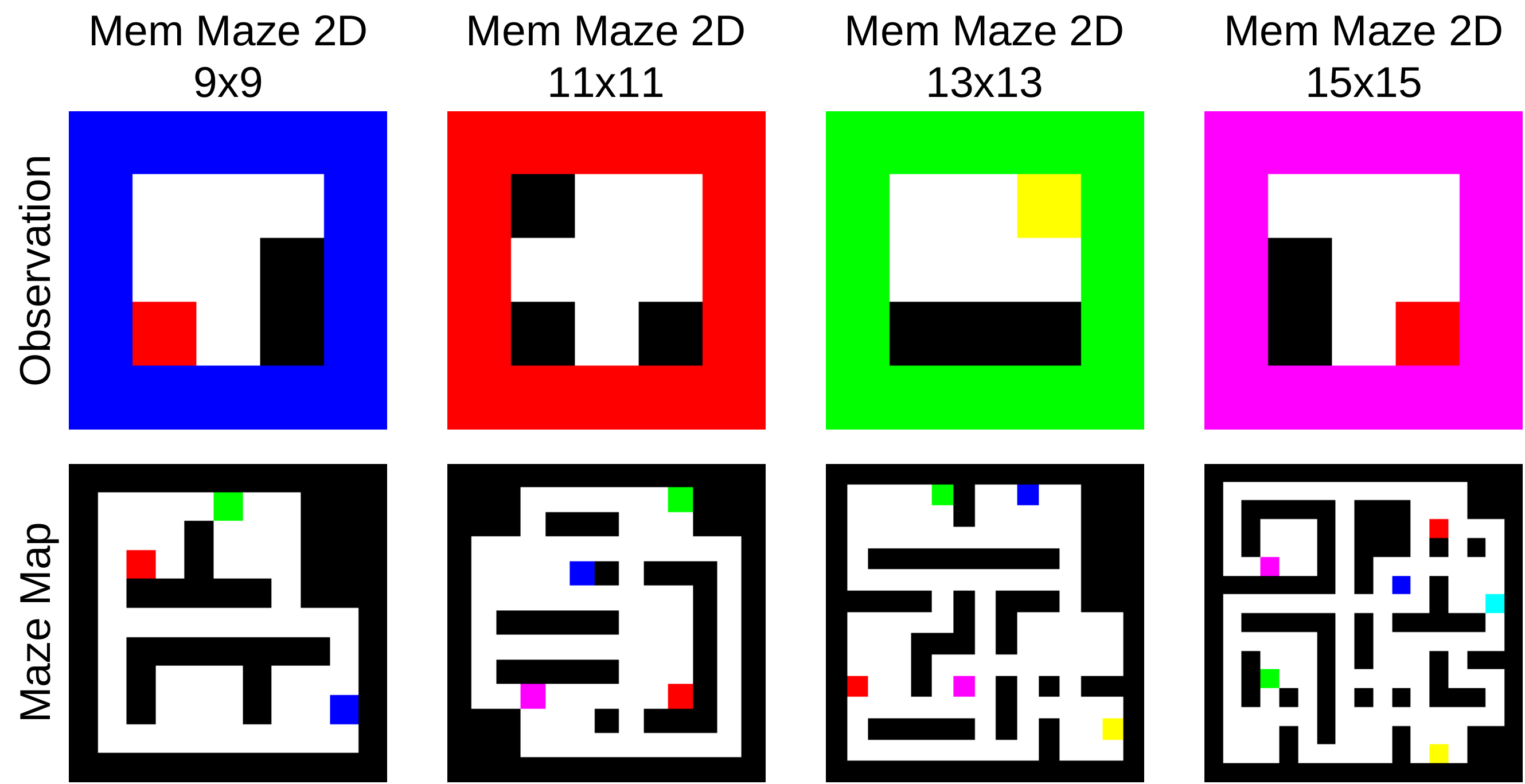}
    \caption{Randomly generated environment map for Memory Maze 2D (9\texttimes 9, 11\texttimes 11, 13\texttimes 13 and 15\texttimes 15). The agent receives a top-down $3 \times 3$ local context centered on its current position and must reach the target color to receive the next reward.}
    \label{figure:memory_maze_2d_map}
\end{figure*}

We design Memory Maze 2D to evaluate the capacity of agents to develop long-term memory by exploring a partially observable environment. Memory Maze 2D serves as a simplified and more efficient version of the Memory Maze environment, where the agent observes a top-down $3 \times 3$ local context centered on its current position. To facilitate goal identification, we augment the observation with colored borders that indicate the target color to be reached for obtaining the next reward. The observation is then flattened and normalized, forming a vector of dimension $75=5\times5\times3$. The agent operates in the maze using a discrete action space consisting of five actions: no-op, move left, move right, move up, and move down. Following~\citep{pasukonis2023evaluating}, we design four maze tasks with varying sizes and difficulty: Memory Maze 2D 9\texttimes9, Memory Maze 2D 11\texttimes11, Memory Maze 2D 13\texttimes13, and Memory Maze 2D 15\texttimes15. Table~\ref{table:memory_maze_2d} shows the parameters associated with each maze size and Figure~\ref{figure:memory_maze_2d_map} depicts randomly generated environment maps.

\begin{table}[ht]
    \caption{Memory Maze 2D environment details.}
    \centering
    \setlength{\tabcolsep}{2pt}
    \hfill \break
    \begin{tabular}{lcccc}
    \toprule
    & \multicolumn{1}{c}{\multirow{2}{*}{\begin{tabular}[c]{@{}c@{}}\textbf{Mem Maze 2D} \\ \textbf{9\texttimes9}\end{tabular}}} & \multicolumn{1}{c}{\multirow{2}{*}{\begin{tabular}[c]{@{}c@{}}\textbf{Mem Maze 2D} \\ \textbf{11\texttimes11}\end{tabular}}} & \multicolumn{1}{c}{\multirow{2}{*}{\begin{tabular}[c]{@{}c@{}}\textbf{Mem Maze 2D} \\ \textbf{13\texttimes13}\end{tabular}}} & \multicolumn{1}{c}{\multirow{2}{*}{\begin{tabular}[c]{@{}c@{}}\textbf{Mem Maze 2D} \\ \textbf{15\texttimes15}\end{tabular}}} \\
    \textbf{Parameter} &  &  &  &  \\
    \midrule
    Number of actions & $5$ & $5$ & $5$ & $5$ \\
    Observation dim & $75$ & $75$ & $75$ & $75$ \\
    Number of colors & $3$ & $4$ & $5$ & $6$ \\
    Number of rooms & $3-4$ & $4-6$ & $5-6$ & $9$ \\
    Room size & $3-5$ & $3-5$ & $3-5$ & $3$ \\
    Episode length & $1000$ & $2000$ & $3000$ & $4000$ \\
    Mean maximum score (Oracle) & $137.2$ & $215.4$  & $276.6$  & $311.4$  \\
    \bottomrule
    \end{tabular}
    \label{table:memory_maze_2d}
\end{table}

\begin{table}[ht]
    \caption{Environment efficiency comparison. Episode average step time measured on a single Intel Core i7-12700 4.9GHz CPU thread and a RTX 3090 GPU.}
    \centering
    \setlength{\tabcolsep}{11pt}
    \hfill \break
    \begin{tabular}{lcccc}
    \toprule
    \textbf{Step time (ms)} & \textbf{9\texttimes9} & \textbf{11\texttimes11} & \textbf{13\texttimes13} & \textbf{15\texttimes15} \\
    \midrule
    Memory Maze 2D & $\hphantom{0}0.16$ & $\hphantom{0}0.16$ & $\hphantom{0}0.16$ & $\hphantom{0}0.16$ \\
    Memory Maze & $47.28$ & $61.82$ & $76.48$ & $90.36$ \\
    \bottomrule
    \end{tabular}
    \label{table:memory_maze_time}
\end{table}

\newpage

\subsection{Algorithm Pseudocode}

\begin{algorithm}[ht]
\caption{TT-MBRL, our proposed algorithm for applying model-based reinforcement learning at test time. At the beginning of each evaluation episode, the agent initializes test-time parameters $(\phi', \theta', \psi')$ from the trained parameters $(\phi, \theta, \psi)$ and creates a dedicated test-time replay buffer $\mathbf{R'}$ to collect episode observations. At each environment step, the world model is updated by sampling from both the test-time buffer $\mathbf{R'}$ and the training buffer $\mathbf{R}$, acting as a regularizer that prevents overfitting to limited early observations. The actor-critic is then updated using the adapted world model, enabling the agent to learn an improved policy.}
\begin{algorithmic}
\Require Trained parameters $(\phi, \theta, \psi)$, training replay buffer $\mathbf{R}$, and environment $env$
\While {not done evaluation}
    \State $(\phi', \theta', \psi') \gets (\phi, \theta, \psi)$
    \Comment{Initialize test-time parameters from training.}
    \State $\mathbf{R'} \gets \{\}$
    \Comment{Initialize empty test-time replay buffer.}
    \State $obs \gets \text{ENV\_RESET}(env)$
    \Comment{Reset environment.}
    \State $\mathbf{R'} \gets \mathbf{R'} \cup \{obs\}$
    \Comment{Store initial observation.}
    \While {not done episode}
        \State $obs \gets \text{ENV\_STEP}(env, \phi', \theta')$
        \Comment{Step environment using current policy.}
        \State $\mathbf{R'} \gets \mathbf{R'} \cup \{obs\}$
        \Comment{Update test-time replay buffer.}
        \State $\phi' \gets \text{TRAIN\_WORLD\_MODEL}(\mathbf{R}, \mathbf{R'})$
        \Comment{Update world model using train-test data.}
        \State $(\theta',~\psi') \gets \text{TRAIN\_ACTOR\_CRITIC}(\phi')$
        \Comment{Update actor-critic using world model.}
    \EndWhile
\EndWhile
\end{algorithmic}
\label{algorithm:pseudocode}
\end{algorithm}

\subsection{Experimental Details and Hyperparameters}
\label{appendix:hyperparameters}

\begin{table*}[!ht]
\caption{World model sizes. N/A refers to the use of a MLP encoder for low-dimensional inputs, and the removal of the AC-CPC objective in the case of TWISTER.}
\begin{subtable}{\textwidth}
    \caption{DreamerV3 model sizes.}
    \centering
    \hfill \break
    \begin{tabular}{lcccc}
    \toprule
    Environment & Mem Maze 2D & Mem Maze & DMC & Atari 100k \\
    \midrule
    CNN Base Dim & N/A & 96 & 32 & 32  \\
    Hidden Size & 512 & 1024 & 512 & 512\\
    GRU Recurrent Units & 512 & 4096 & 512 & 512\\
    MLP Layers & 2 & 5 & 2 & 2\\
    \midrule
    Number Params & 8.7M & 200.7M & 17.9M & 17.9M \\
    \bottomrule
    \end{tabular}
\end{subtable}
\\ \\
\begin{subtable}{0.5\textwidth}
    \setlength{\tabcolsep}{2pt}
    \caption{TWISTER model sizes.}
    \centering
    \hfill \break
    \begin{tabular}{lcc}
    \toprule
    Environment & Mem Maze 2D & Mem Maze \\
    \midrule
    CNN Base Dim & N/A & 48 \\
    Hidden Size & 512 & 640 \\
    Transformer Blocks & 4 & 5 \\
    Feed-Forward Ratio & 2 & 3 \\
    Number of Heads & 8 & 10 \\
    Dropout Prob & 0.1 & 0.1 \\
    MLP Layers & 2 & 3 \\
    Attention Context & 256 & 256 \\
    AC-CPC Steps & N/A & 10 \\
    \midrule
    Number params & 15.9M & 58.8M \\
    \bottomrule
    \end{tabular}
\end{subtable}
\hspace{\fill}
\begin{subtable}{0.5\textwidth}
    \setlength{\tabcolsep}{2pt}
    \caption{R2I model sizes.}
    \centering
    \hfill \break
    \begin{tabular}{lcc}
    \toprule
    Environment & Mem Maze 2D & Mem Maze \\
    \midrule
    CNN Base Dim & N/A & 48 \\
    Hidden Size & 1024 & 2048 \\
    SSM Layers & 3 & 5 \\
    SSM State Dim & 128 & 512 \\
    SSM Output Dim & 1024 & 1024 \\
    MLP Layers & 2 & 4 \\
    \midrule
    Number params & 11.6M & 92.8M \\
    \bottomrule
    \end{tabular}
\end{subtable}

\label{table:wm_params}
\end{table*}

\newpage

\begin{table*}[!ht]
\caption{DreamerV3 hyperparameters.}
\setlength{\tabcolsep}{20pt}
\renewcommand{\arraystretch}{1.25}
\centering
\hfill \break
\begin{tabular}{lcc}
\toprule
Hyperparameter & Memory Maze 2D & Memory Maze \\ 
\midrule
General & \\
Batch Size (B) & 64 & 16 \\
Sequence Length (T) & 64 & 64 \\
Environment Instances & 8 & 32 \\
Policy Step per Grad Step & 64 & 64 \\
Replay Buffer Capacity & 10M & 10M \\
TT-MBRL Policy Step / Grad Step & 1 & 1 \\
TT-MBRL Batch Size & 32 & 8 \\
TT-MBRL Train/Test Ratio & 50/50 & 50/50 \\
\midrule
World Model & \\
Prediction Loss Scale ($\beta_{\text{pred}}$) & 1.0 & 1.0 \\
Dynamics Loss Scale ($\beta_{\text{dyn}}$) & 0.5 & 0.5 \\
Representation Loss Scale ($\beta_{\text{rep}}$) & 0.1 & 0.1\\
Stochastic State Features & 32 & 32\\
Classes per Feature & 32 & 32\\
Learning Rate & $10^{-4}$ & $10^{-4}$\\
Gradient Clipping & 1000 & 1000 \\
\midrule
Actor-Critic & \\
Imagination Horizon ($H$) & 15 & 15\\
Return Discount ($\gamma$) & 0.997 & 0.997\\
Return Lambda ($\lambda$) & 0.95 & 0.95\\
Critic EMA Decay & 0.98 & 0.98 \\
Return Normalization Momentum & 0.99 & 0.99 \\
Actor Entropy Scale ($\eta$) & $3 \cdot 10^{-4}$ & $3 \cdot 10^{-4}$\\
Learning Rate & $3 \cdot 10^{-5}$ & $3 \cdot 10^{-5}$\\
Gradient Clipping & 100 & 100 \\
\bottomrule
\end{tabular}
\label{table:hyperparams_dv3}
\end{table*}

\newpage

\begin{table*}[!ht]
\caption{TWISTER hyperparameters.}
\setlength{\tabcolsep}{20pt}
\renewcommand{\arraystretch}{1.25}
\centering
\hfill \break
\begin{tabular}{lcc}
\toprule
Hyperparameter & Memory Maze 2D & Memory Maze \\ 
\midrule
General & \\
Batch Size (B) & 4 & 4 \\
Sequence Length (T) & 1024 & 1024 \\
Environment Instances & 8 & 32 \\
Policy Step per Grad Step & 64 & 64 \\
Replay Buffer Capacity & 10M & 10M \\
TT-MBRL Policy Step / Grad Step & 1 & 1 \\
TT-MBRL Batch Size & 2 & 2 \\
TT-MBRL Train/Test Ratio & 50/50 & 50/50 \\
\midrule
World Model & \\
Prediction Loss Scale ($\beta_{\text{pred}}$) & 1.0 & 1.0 \\
Dynamics Loss Scale ($\beta_{\text{dyn}}$) & 0.5 & 0.5 \\
Representation Loss Scale ($\beta_{\text{rep}}$) & 0.1 & 0.1\\
Stochastic State Features & 32 & 32\\
Classes per Feature & 32 & 32\\
Learning Rate & $10^{-4}$ & $10^{-4}$\\
Gradient Clipping & 1000 & 1000 \\
\midrule
Actor-Critic & \\
Imagination Horizon ($H$) & 15 & 15\\
Return Discount ($\gamma$) & 0.997 & 0.997\\
Return Lambda ($\lambda$) & 0.95 & 0.95\\
Critic EMA Decay & 0.98 & 0.98 \\
Return Normalization Momentum & 0.99 & 0.99 \\
Actor Entropy Scale ($\eta$) & $3 \cdot 10^{-4}$ & $3 \cdot 10^{-4}$\\
Learning Rate & $3 \cdot 10^{-5}$ & $3 \cdot 10^{-5}$\\
Gradient Clipping & 100 & 100 \\
\bottomrule
\end{tabular}
\label{table:hyperparams_twister}
\end{table*}

\newpage

\begin{table*}[!ht]
\caption{R2I hyperparameters.}
\setlength{\tabcolsep}{20pt}
\renewcommand{\arraystretch}{1.25}
\centering
\hfill \break
\begin{tabular}{lcc}
\toprule
Hyperparameter & Memory Maze 2D & Memory Maze \\ 
\midrule
General & \\
Batch Size (B) & 4 & 4 \\
Sequence Length (T) & 1024 & 1024 \\
Environment Instances & 8 & 40 \\
Policy Step per Grad Step & 64 & 64 \\
Replay Buffer Capacity & 10M & 10M \\
TT-MBRL Policy Step / Grad Step & 1 & 1 \\
TT-MBRL Batch Size & 2 & 2 \\
TT-MBRL Train/Test Ratio & 50/50 & 50/50 \\
\midrule
World Model & \\
Prediction Loss Scale ($\beta_{\text{pred}}$) & 1.0 & 1.0 \\
Dynamics Loss Scale ($\beta_{\text{dyn}}$) & 0.8 & 0.8 \\
Representation Loss Scale ($\beta_{\text{rep}}$) & 0.2 & 0.2 \\
Stochastic State Features & 32 & 32\\
Classes per Feature & 32 & 32\\
Learning Rate & $3 \cdot 10^{-4}$ & $3 \cdot 10^{-4}$\\
Gradient Clipping & 200 & 200 \\
\midrule
Actor-Critic & \\
Imagination Horizon ($H$) & 15 & 15\\
Return Discount ($\gamma$) & 0.997 & 0.997\\
Return Lambda ($\lambda$) & 0.95 & 0.95\\
Critic EMA Decay & 0.98 & 0.98 \\
Return Normalization Momentum & 0.99 & 0.99 \\
Actor Entropy Scale ($\eta$) & $10^{-3}$ & $10^{-3}$\\
Learning Rate & $10^{-4}$ & $10^{-4}$\\
Gradient Clipping & 200 & 200 \\
\bottomrule
\end{tabular}
\label{table:hyperparams_r2i}
\end{table*}

\newpage

\subsection{Training Curves}

\begin{figure}[!ht]
        \centering
        \includegraphics[width=\linewidth]{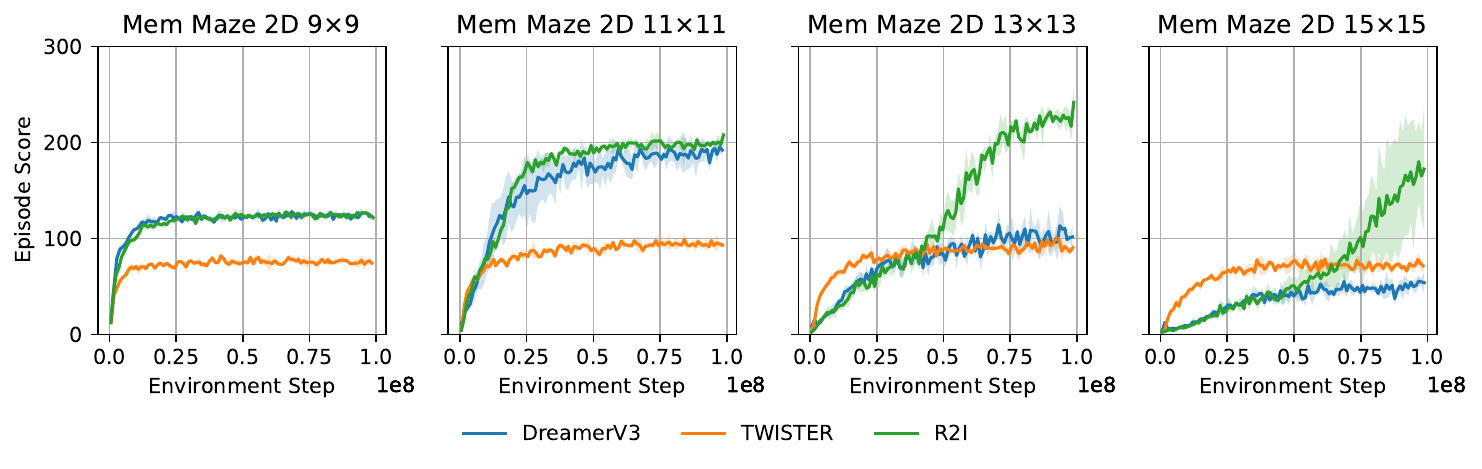}
        \caption{Memory Maze 2D training curves (100M environment steps).}
        \label{appendix:training_curves_2d}
\end{figure}

\begin{figure}[!ht]
        \centering
        \includegraphics[width=\linewidth]{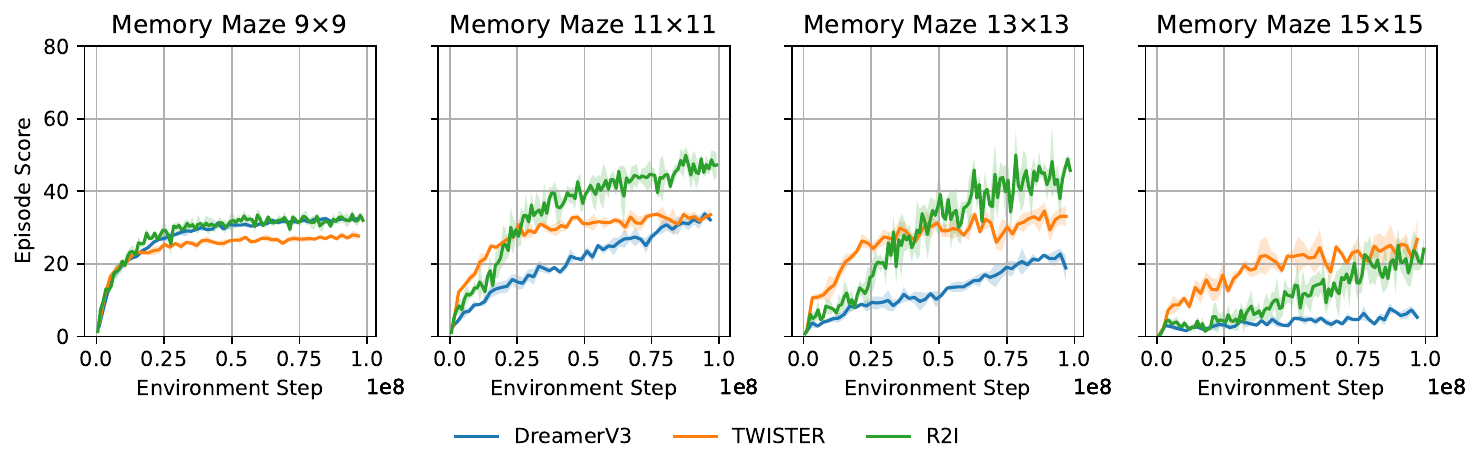}
        \caption{Memory Maze training curves (100M environment steps).}
        \label{appendix:training_curves_3d_100M}
\end{figure}

\begin{figure}[!ht]
        \centering
        \includegraphics[width=\linewidth]{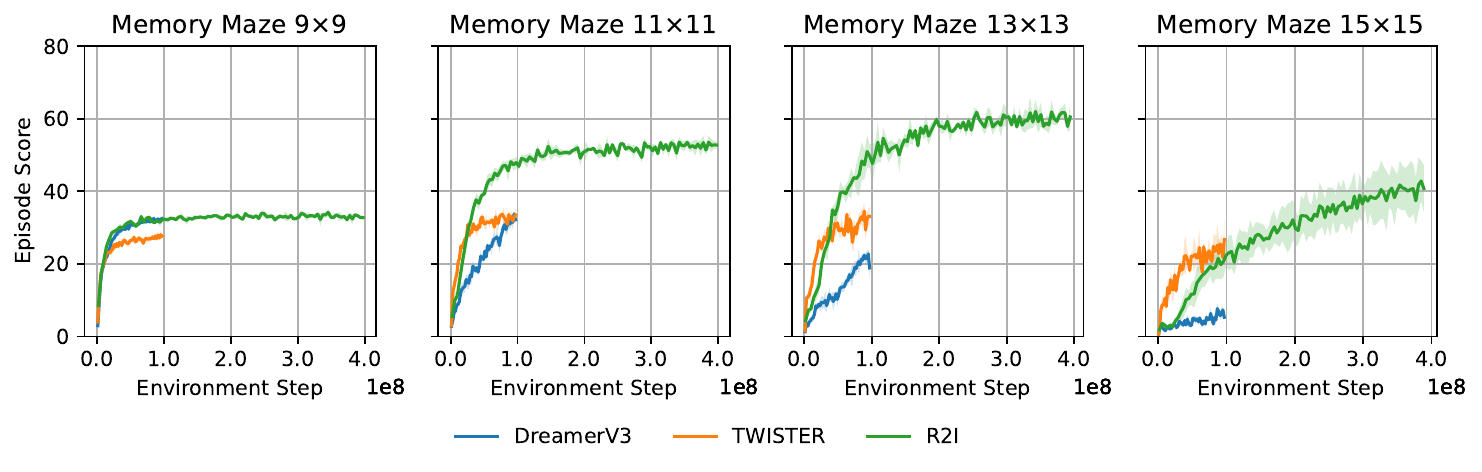}
        \caption{Memory Maze training curves (400M environment steps).}
        \label{appendix:training_curves_3d}
\end{figure}

\newpage

\subsection{Evaluation Histograms}
\label{appendix:histograms}

\begin{figure*}[ht]
    \centering
    \includegraphics[width=\linewidth]{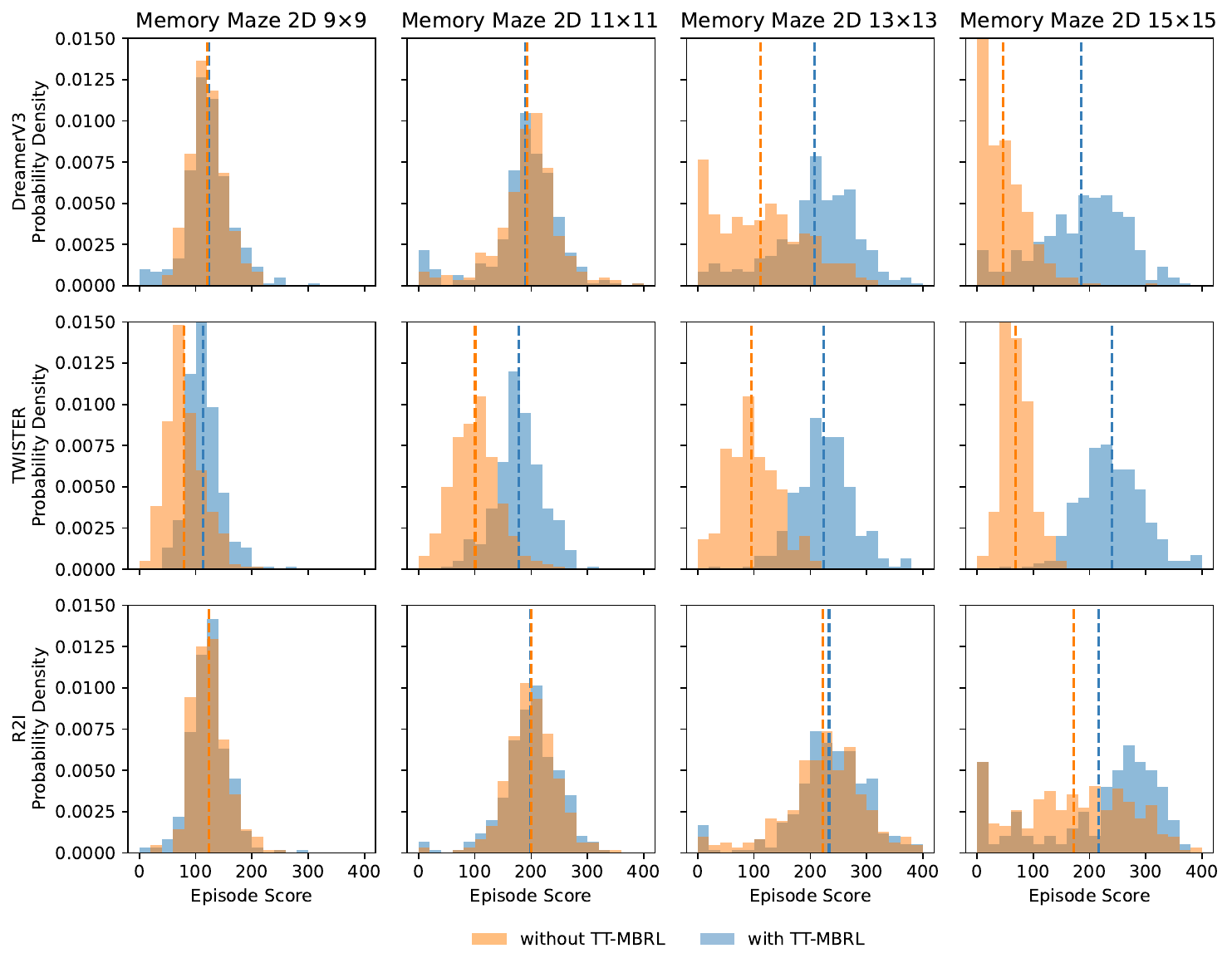}
    \caption{Evaluation episode return distribution on Memory Maze 2D (100M environment steps). We compare the impact of TT-MBRL on episode return for DreamerV3, TWISTER and R2I.}
    \label{figure:hist_memory_maze_2d}
\end{figure*}

\begin{figure*}[ht]
    \centering
    \includegraphics[width=1.0\linewidth]{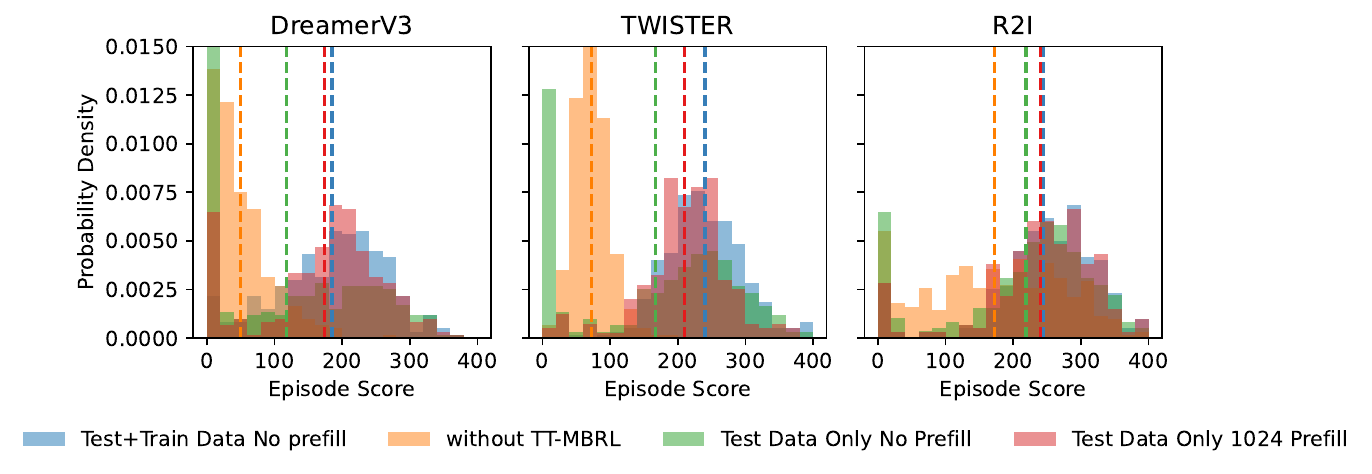}
    \caption{Evaluation episode return distribution on Memory Maze 2D 15\texttimes 15. We compare the impact of training trajectories regularization and replay buffer prefill on TT-MBRL efficacy.}
    \label{figure:study_test_train_hist}
\end{figure*}

\newpage

\begin{figure*}[ht]
    \centering
    \includegraphics[width=\linewidth]{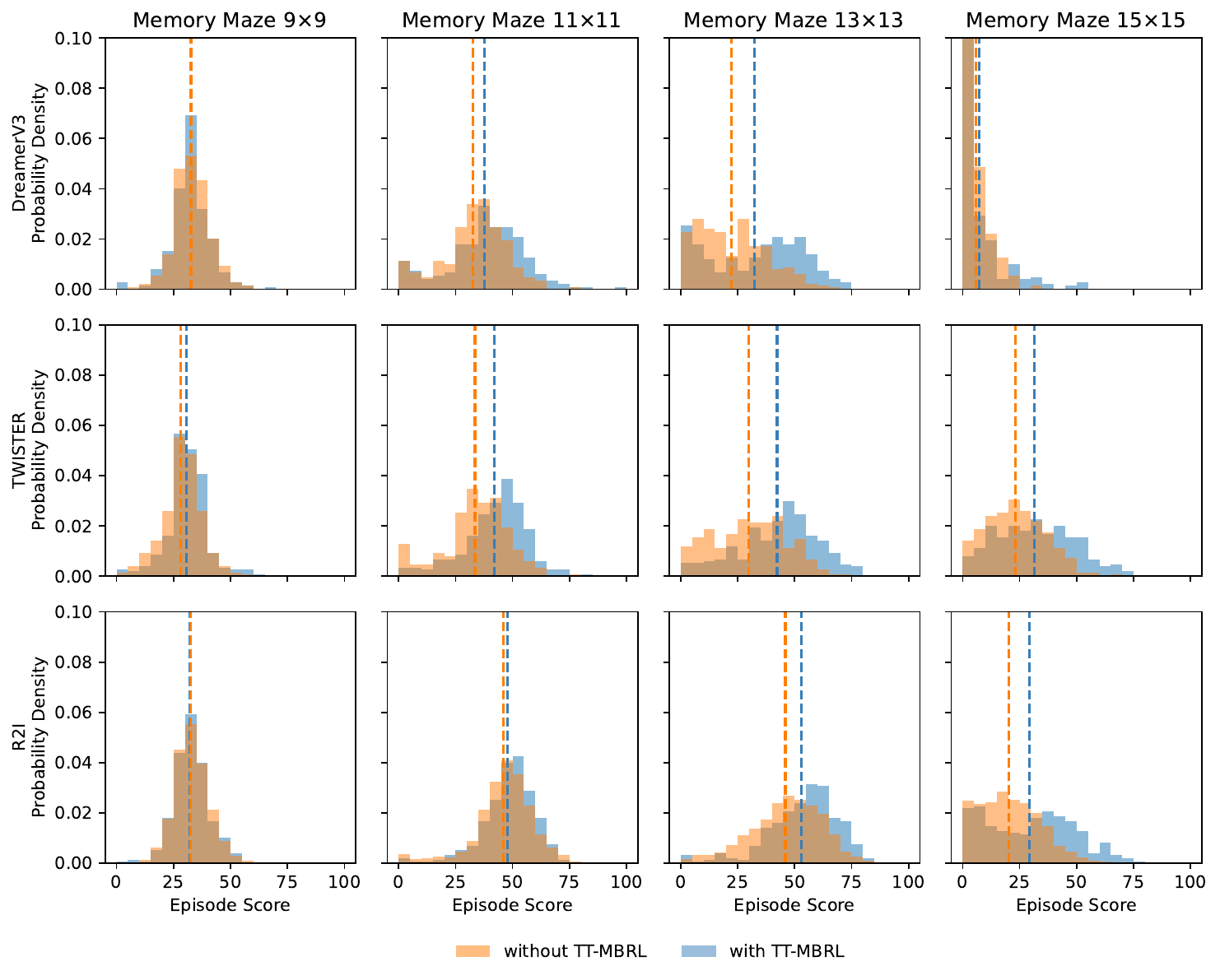}
    \caption{Evaluation episode return distribution on Memory Maze (100M environment steps). We show the impact of TT-MBRL on episode return for DreamerV3, TWISTER and R2I.}
    \label{figure:hist_memory_maze_3d_100M}
\end{figure*}

\begin{figure*}[ht]
    \centering
    \includegraphics[width=\linewidth]{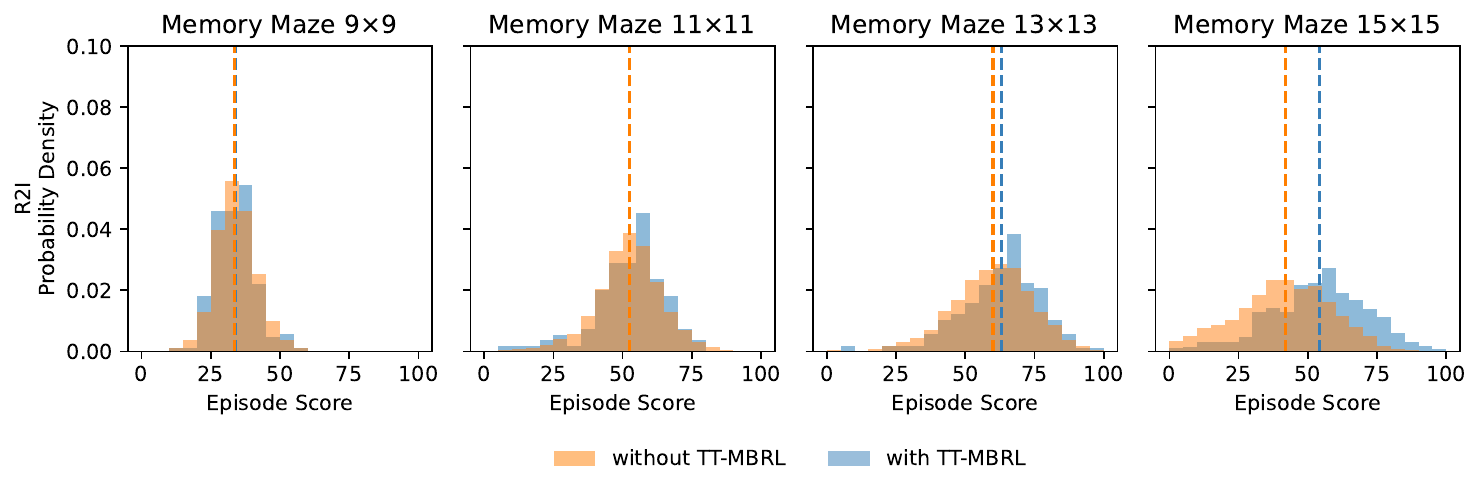}
    \caption{Evaluation episode return distribution on Memory Maze (400M environment steps). We show the impact of TT-MBRL on episode return for R2I.}
    \label{figure:hist_memory_maze_3d}
\end{figure*}

\newpage

\begin{figure*}[ht]
    \centering
    \includegraphics[width=\linewidth]{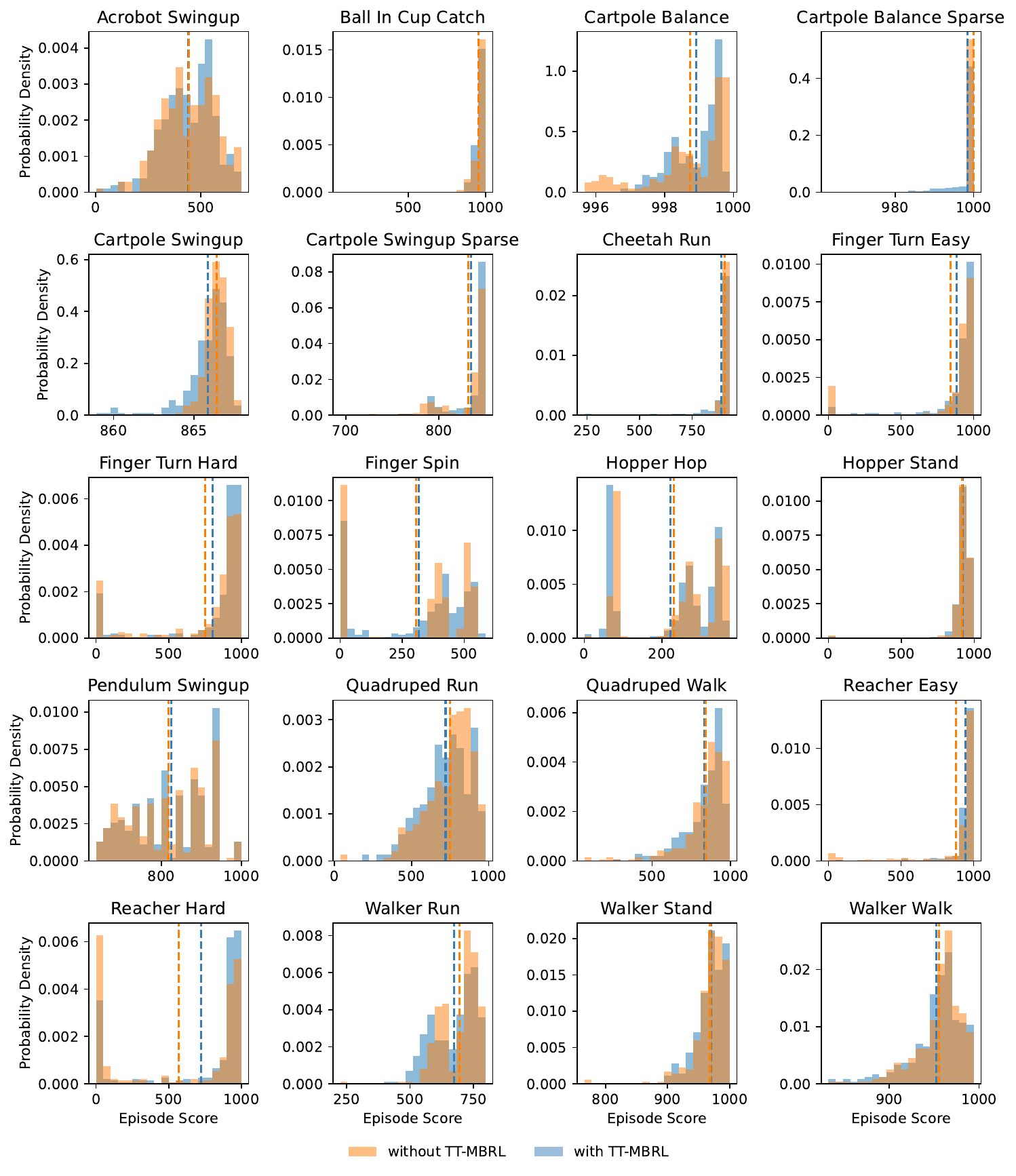}
    \caption{Evaluation episode return distribution on the DeepMind Control Suite (DMC). We show the impact of TT-MBRL on episode return for DreamerV3.}
    \label{figure:hist_dmc}
\end{figure*}

\newpage

\begin{figure*}[ht]
    \centering
    \includegraphics[width=0.79\linewidth]{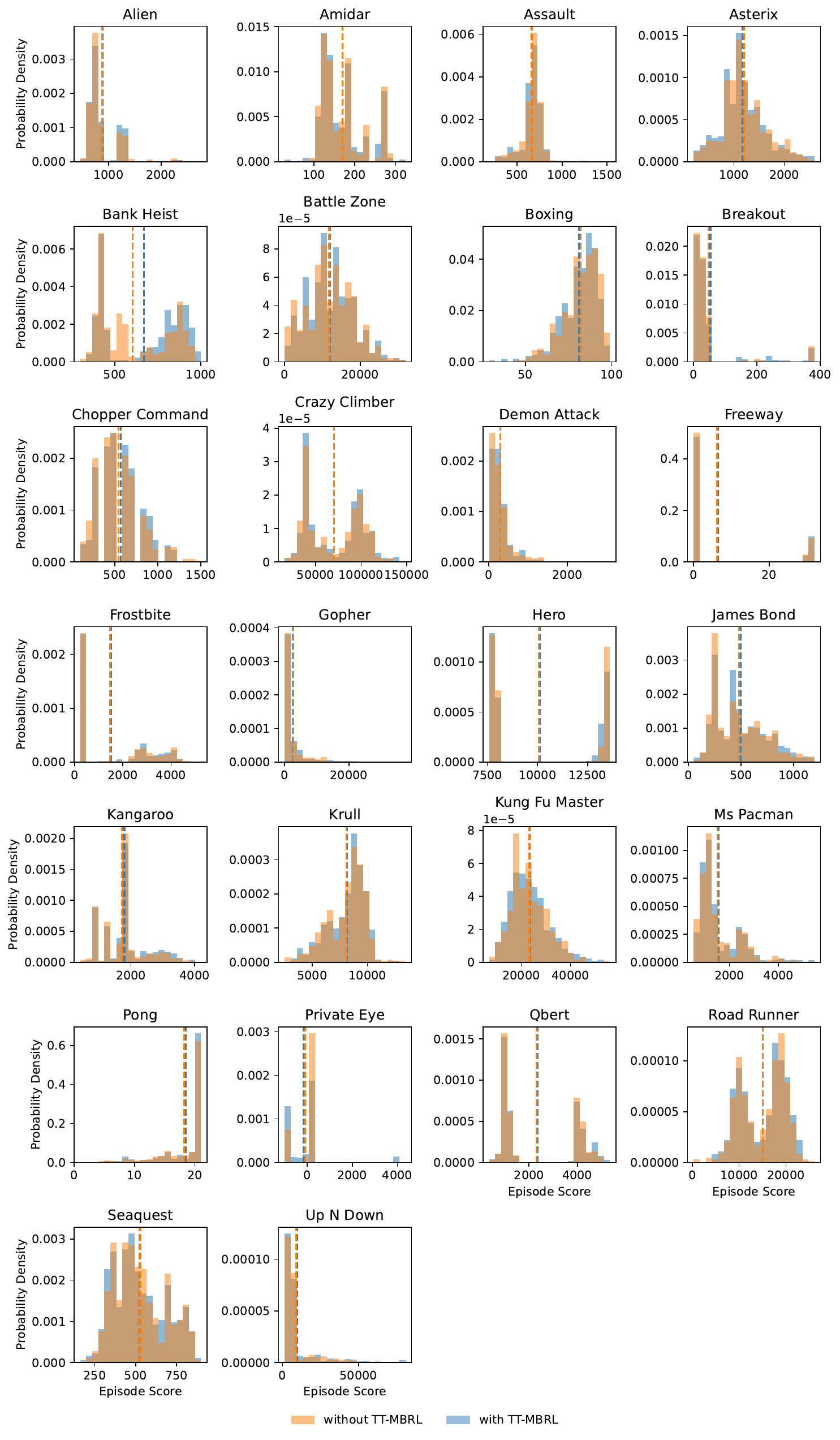}
    \caption{Evaluation episode return distribution on the Atari 100k benchmark. We show the impact of TT-MBRL on episode return for DreamerV3.}
    \label{figure:hist_atari100k}
\end{figure*}

\newpage

\subsection{TT-MBRL Relative Improvement}

\begin{figure*}[ht]
    \centering
    \includegraphics[width=1.0\linewidth]{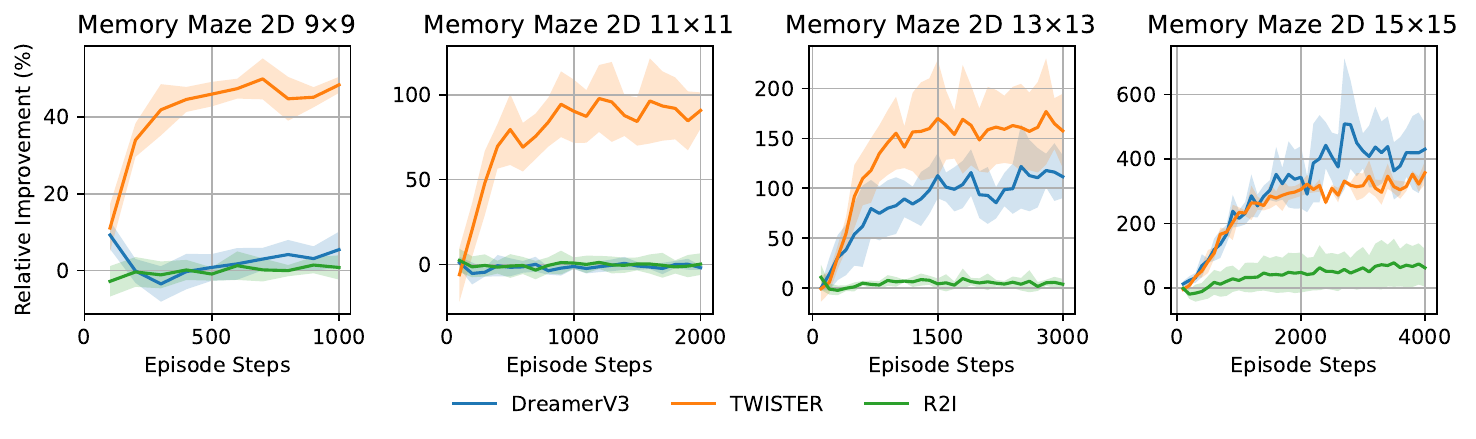}
    \caption{Relative improvement in obtained rewards brought by TT-MBRL on Memory Maze 2D.}
    \label{figure:rel_impr_2d}
\end{figure*}

\begin{figure*}[ht]
    \centering
    \includegraphics[width=1.0\linewidth]{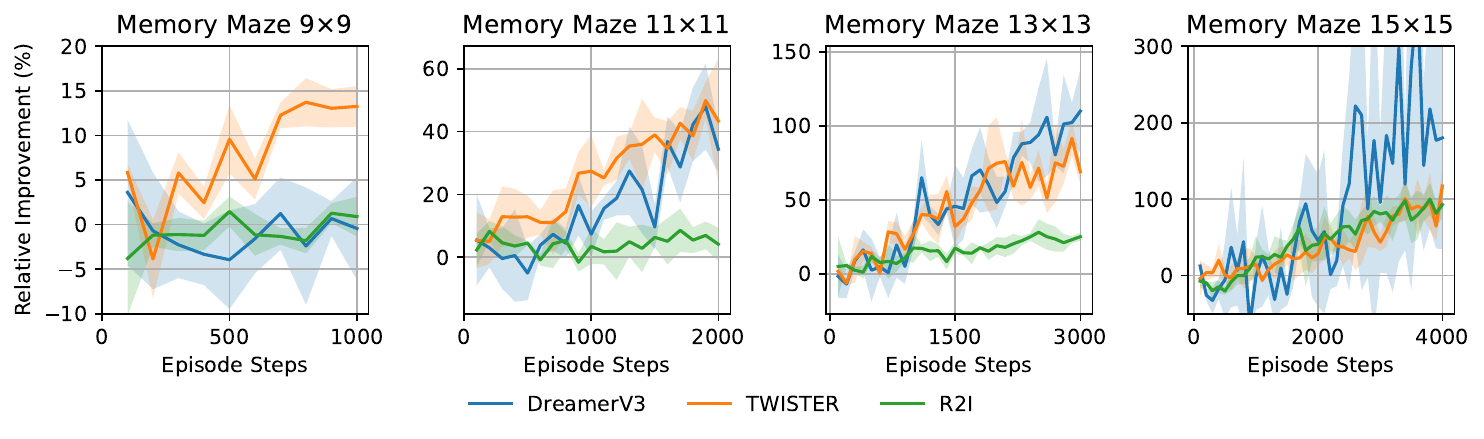}
    \caption{Relative improvement in obtained rewards brought by TT-MBRL on Memory Maze (100M environment steps).}
    \label{figure:rel_impr_3d_100M}
\end{figure*}

\begin{figure*}[ht]
    \centering
    \includegraphics[width=1.0\linewidth]{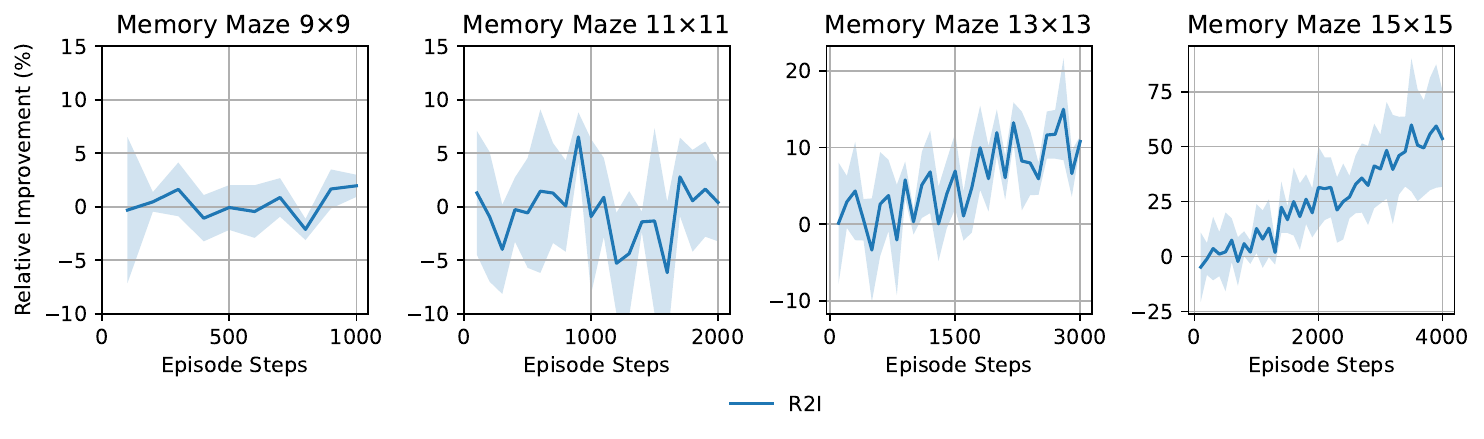}
    \caption{Relative improvement in obtained rewards brought by TT-MBRL on Memory Maze (400M environment steps).}
    \label{figure:rel_impr_3d}
\end{figure*}

\newpage

\subsection{World Model Prediction Metrics}

\begin{figure*}[ht]
    \centering
    \includegraphics[width=1.0\linewidth]{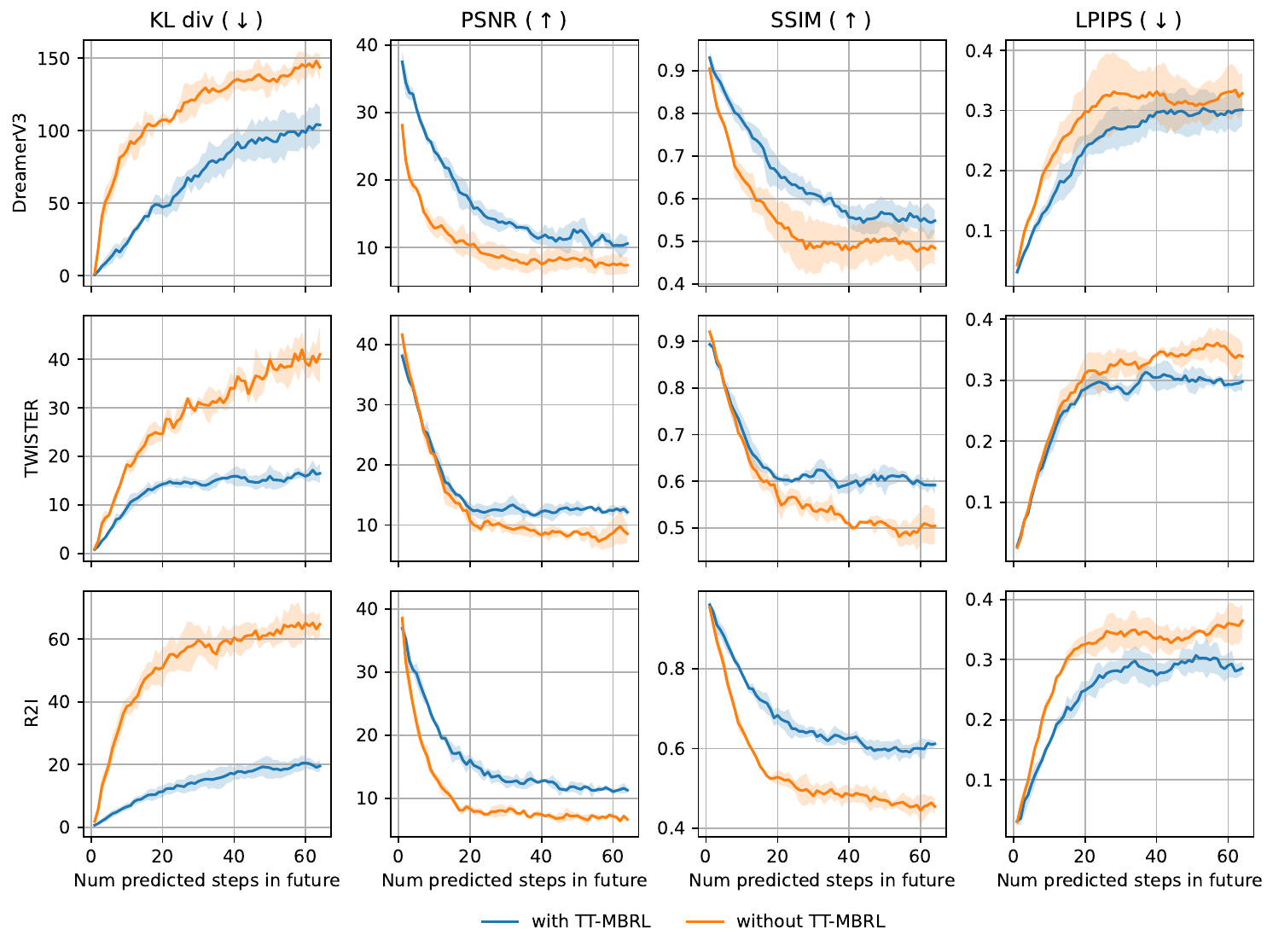}
    \caption{DreamerV3, TWISTER and R2I world model prediction metrics on Memory Maze 2D 15\texttimes 15.}
    \label{figure:metrics_2d}
\end{figure*}

\newpage

\begin{figure*}[ht]
    \centering
    \includegraphics[width=1.0\linewidth]{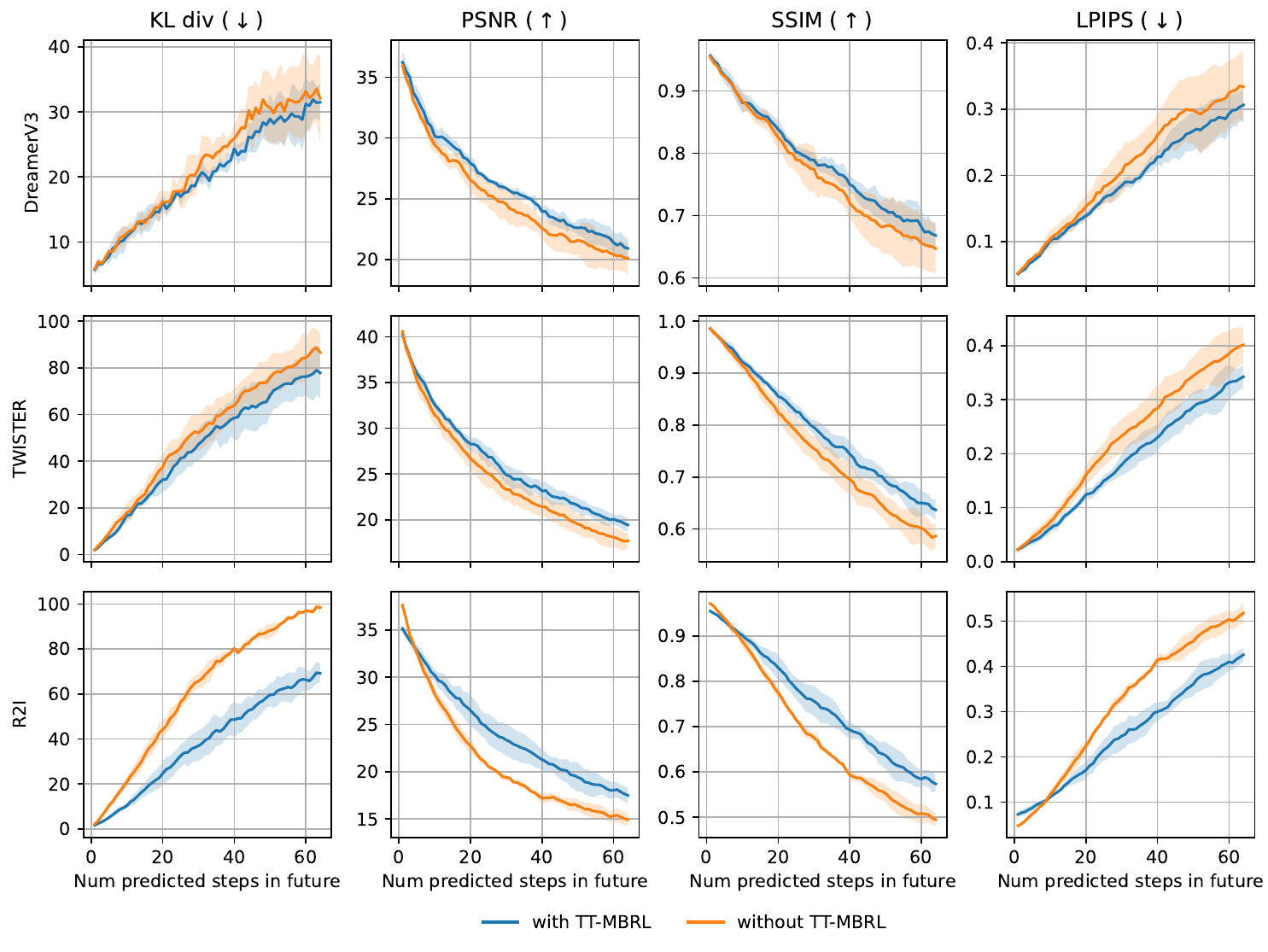}
    \caption{DreamerV3, TWISTER and R2I world model prediction metrics on Memory Maze 15\texttimes 15 (100M environment steps).}
    \label{figure:metrics_3d_100M}
\end{figure*}

\begin{figure*}[ht]
    \centering
    \includegraphics[width=1.0\linewidth]{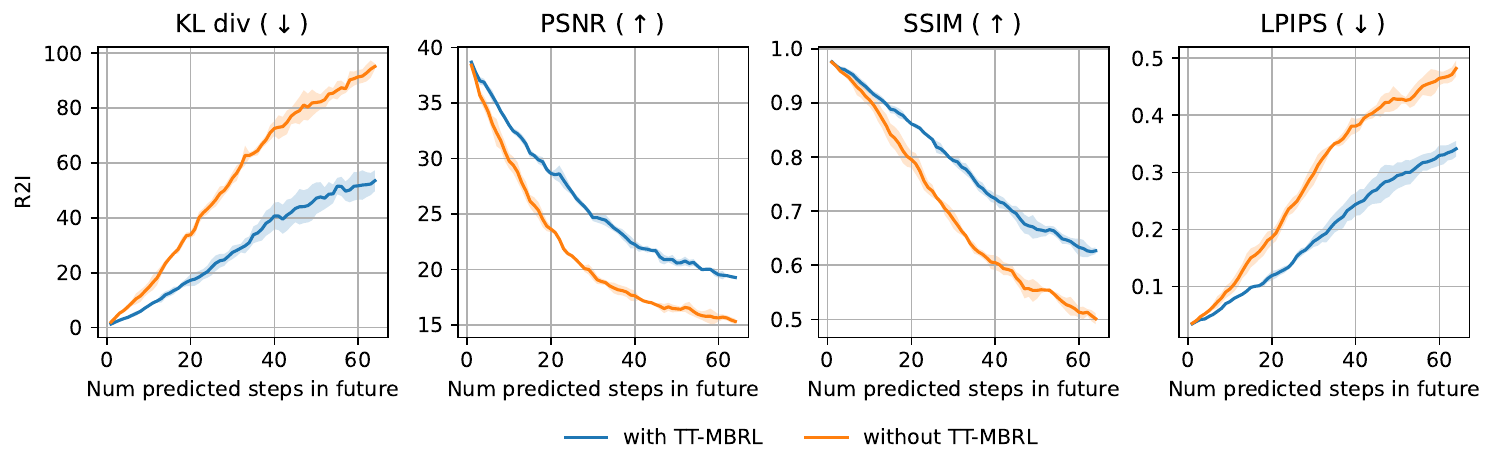}
    \caption{R2I world model prediction metrics on Memory Maze 15\texttimes 15 (400M environment steps).}
    \label{figure:metrics_3d}
\end{figure*}

\newpage

\newpage

\subsection{World Model Predictions}

\begin{figure*}[ht]
    \centering
    \includegraphics[width=0.85\linewidth]{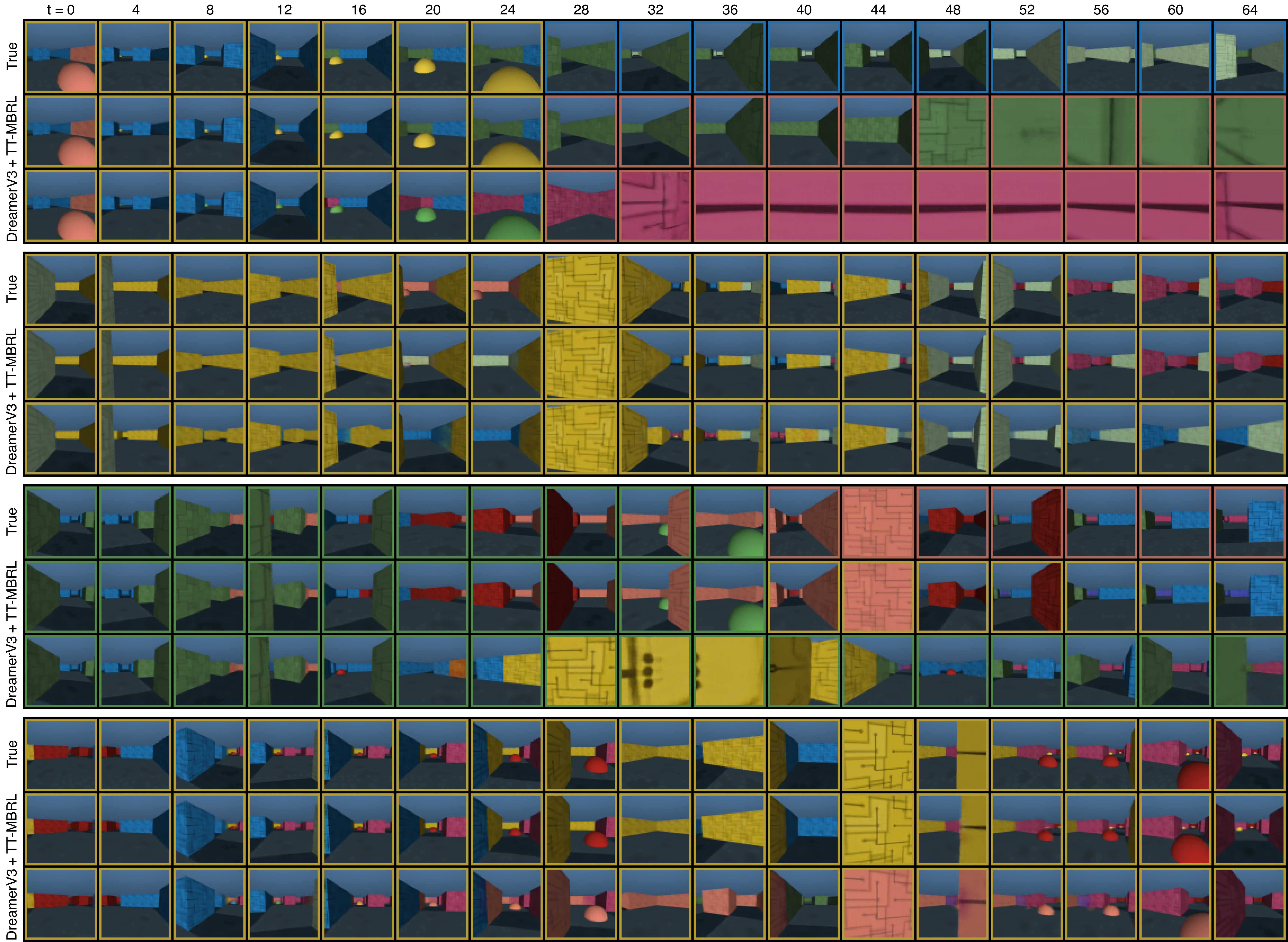}
    \caption{DreamerV3 world model predictions on Memory Maze.}
    \label{figure:traj_memory_maze_dv3}
\end{figure*}

\begin{figure*}[ht]
    \centering
    \includegraphics[width=0.85\linewidth]{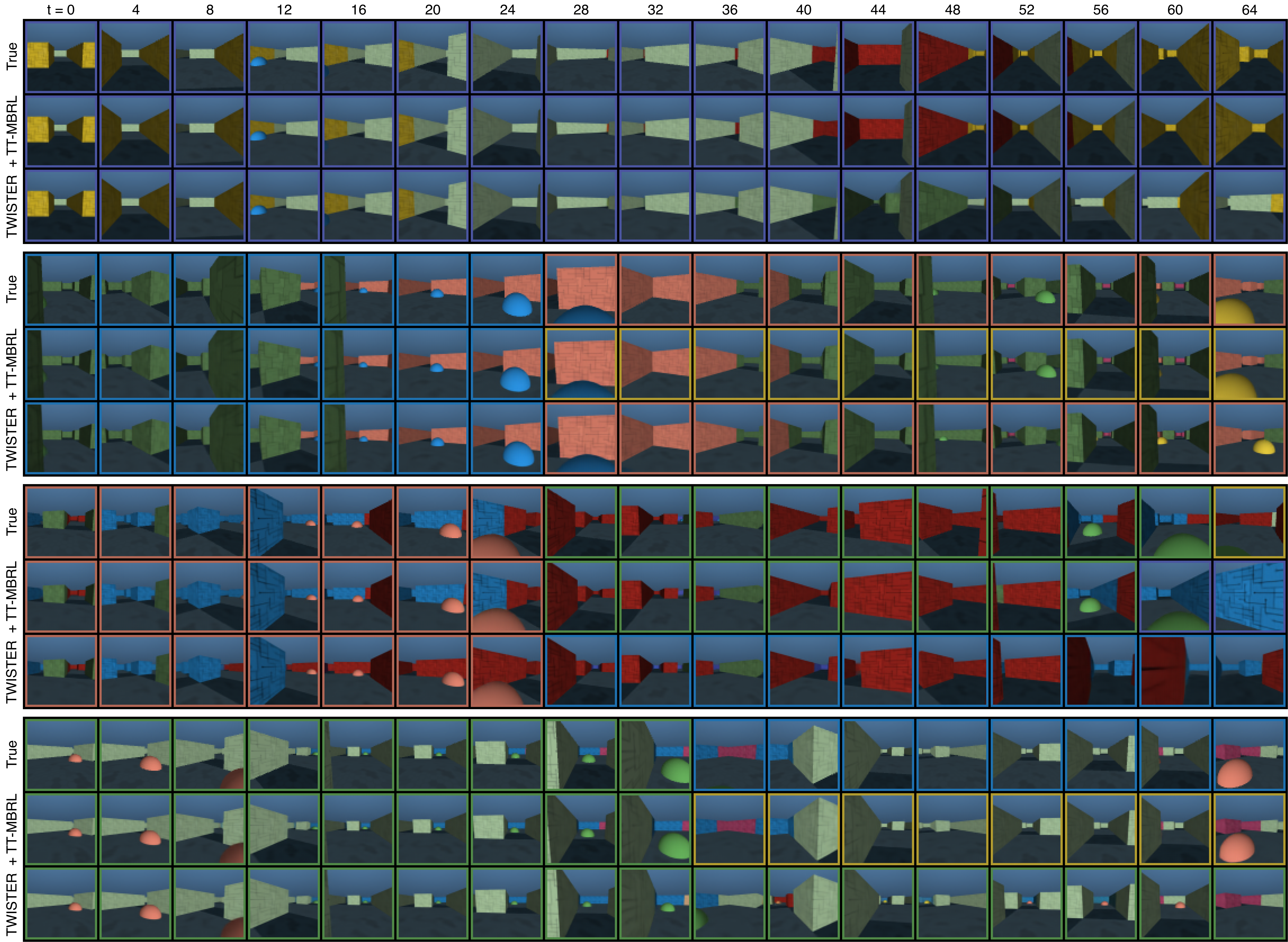}
    \caption{TWISTER world model predictions on Memory Maze.}
    \label{figure:traj_memory_maze_twister}
\end{figure*}

\newpage

\begin{figure*}[ht]
    \centering
    \includegraphics[width=0.85\linewidth]{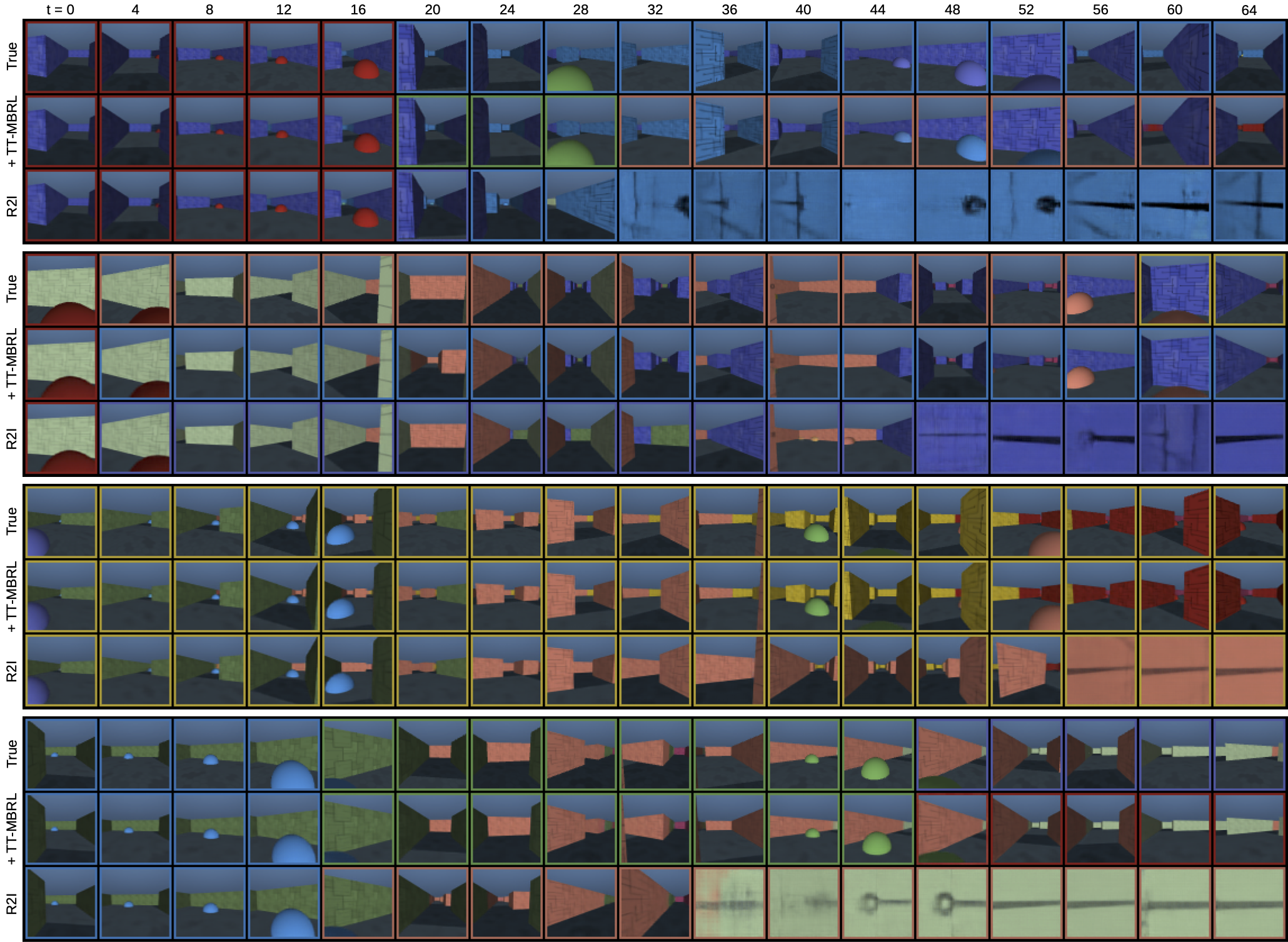}
    \caption{R2I world model predictions on Memory Maze.}
    \label{figure:traj_memory_maze_r2i}
\end{figure*}

\begin{figure*}[ht]
    \centering
    \includegraphics[width=0.85\linewidth]{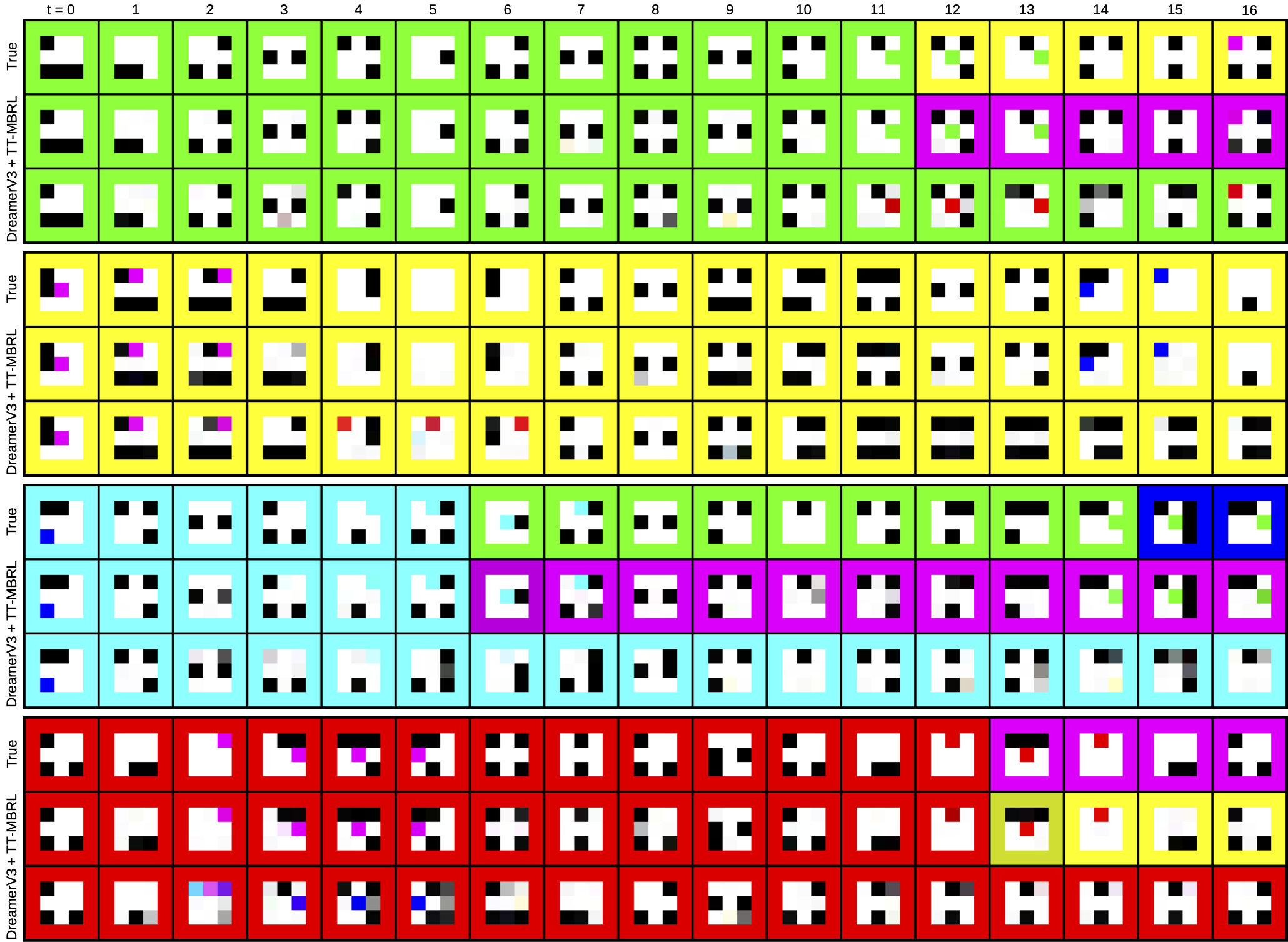}
    \caption{DreamerV3 world model predictions on Memory Maze 2D.}
    \label{figure:traj_memory_maze_2d}
\end{figure*}

\newpage

\begin{figure*}[ht]
    \centering
    \includegraphics[width=0.85\linewidth]{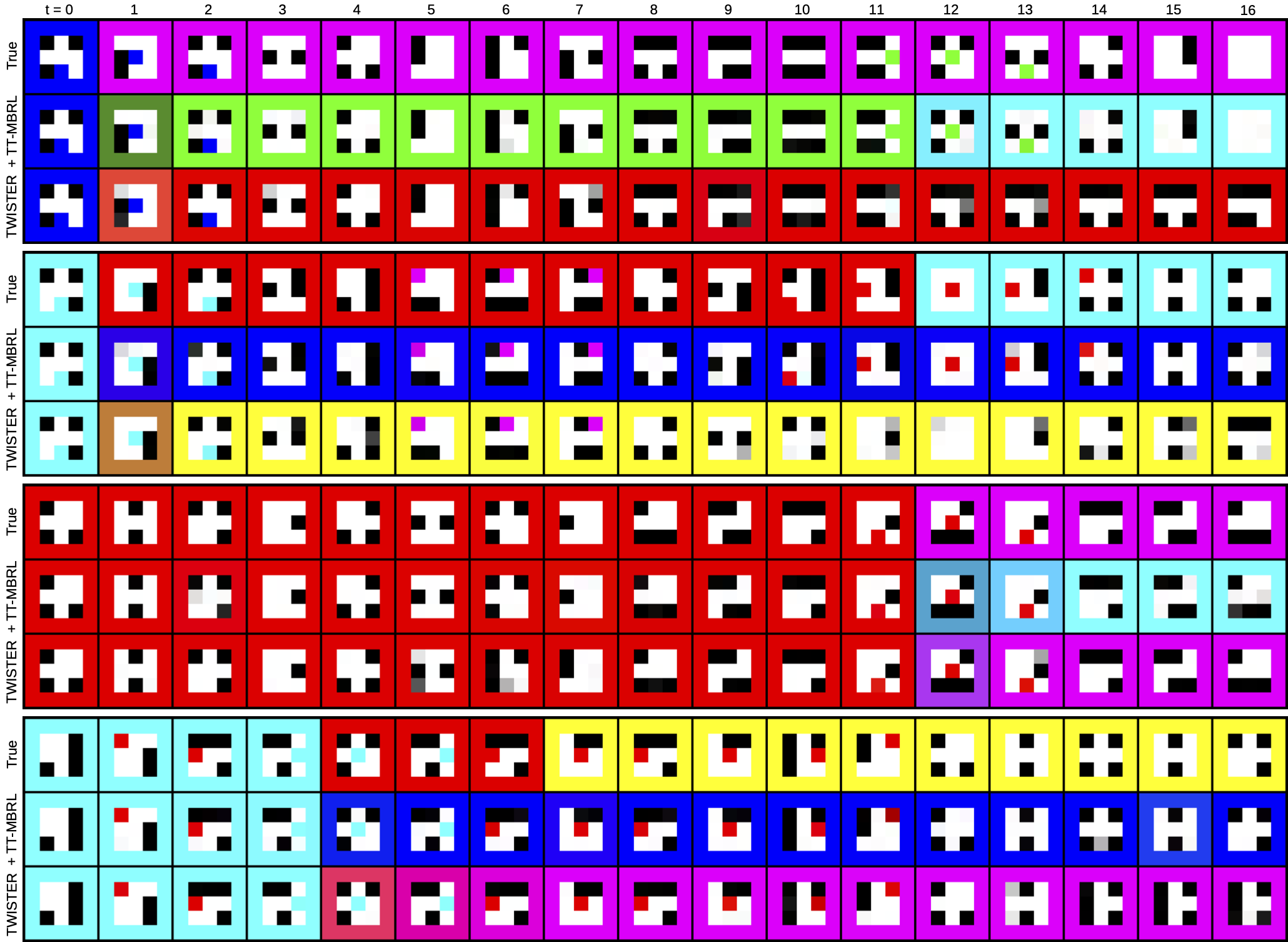}
    \caption{TWISTER world model predictions on Memory Maze 2D.}
    \label{figure:traj_memory_maze_2d_twister}
\end{figure*}

\begin{figure*}[ht]
    \centering
    \includegraphics[width=0.85\linewidth]{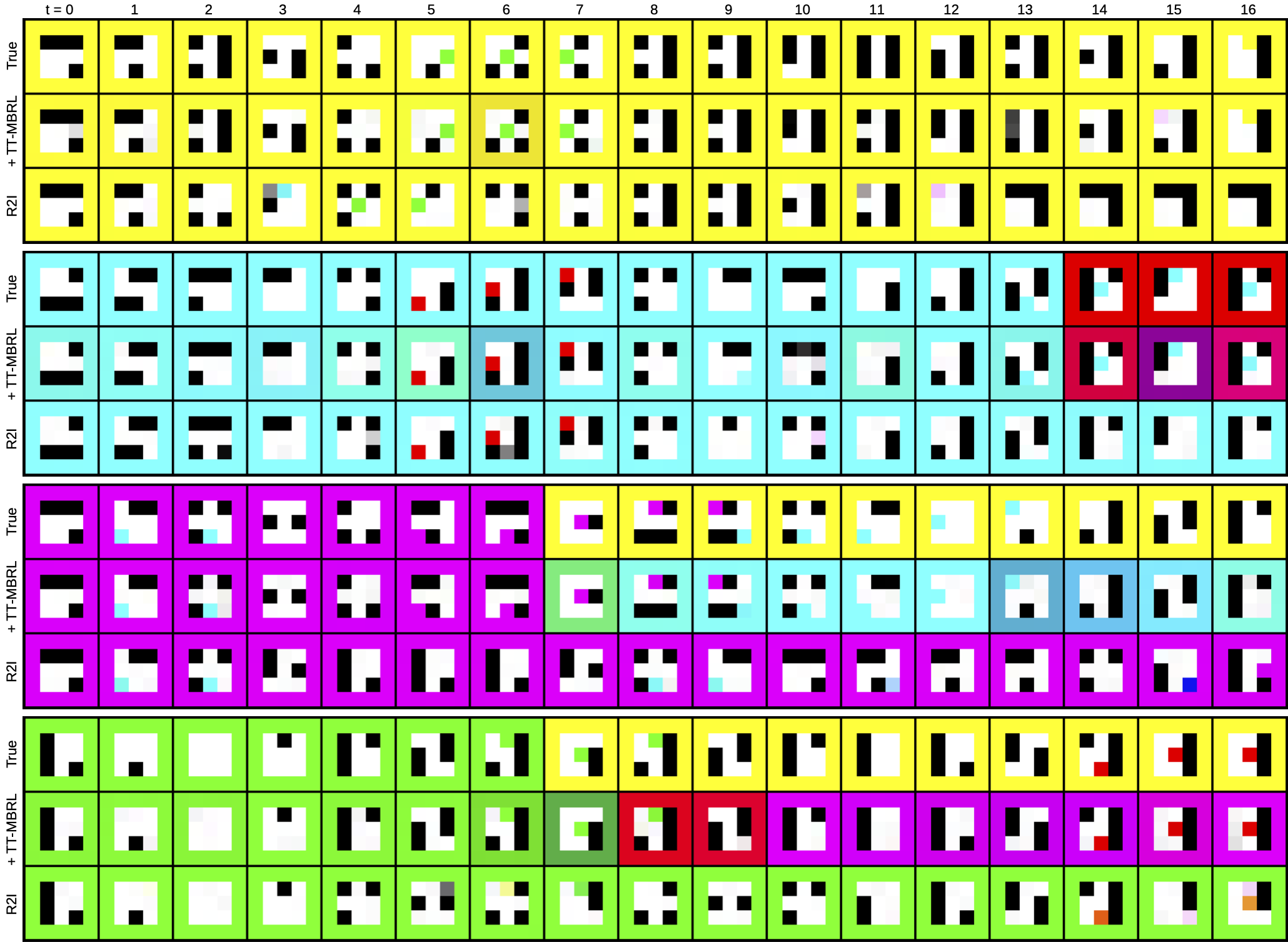}
    \caption{R2I world model predictions on Memory Maze 2D.}
    \label{figure:traj_memory_maze_2d_r2i}
\end{figure*}

%% file: sec/2_related_works.tex
\subsection{Related Works}

\subsubsection{Model-based Reinforcement Learning}

Model-based reinforcement learning approaches use a model of the environment to simulate agent trajectories, enabling better generalization, sample efficiency, and decision-making through planning. Following the success of deep neural networks for learning function approximations~\citep{lecun2015deep}, researchers proposed to learn world models using gradient backpropagation. One of the first approaches to apply model-based reinforcement learning to high-dimensional visual observations is SimPLe~\citep{kaiser2019model}. The authors proposed to learn a predictive world model for Atari games directly in pixel space using a convolutional autoencoder. SimPLe's model predicts future observations and rewards conditioned on past frames and actions, enabling the training of a Proximal Policy Optimization (PPO) agent~\citep{schulman2017proximal} using simulated rollouts.

PlaNet~\citep{hafner2019learning} subsequently introduced Recurrent State-Space Models (RSSMs), combining deterministic recurrent dynamics implemented with Gated Recurrent Units (GRUs)~\citep{cho2014learning} and stochastic latent variables to learn compact world models in latent space. The agent learns hidden representations by encoding observations using a convolutional variational autoencoder (VAE)~\citep{kingma2014auto} while predicting future latent states and rewards, enabling planning via model predictive control. Building on this framework, Dreamer~\citep{hafner2019dream} improved performance by learning a policy and value function directly within the learned latent space. DreamerV2~\citep{hafner2020mastering} extended this approach to Atari games by introducing categorical latent variables trained with straight-through gradients~\citep{bengio2013estimating}, improving representational capacity and stability in discrete domains. More recently, DreamerV3~\citep{hafner2023mastering} demonstrated that a single architecture with shared hyperparameters can scale across diverse domains, from continuous control to Atari. This was achieved through a set of architectural and optimization refinements, including symlog-scaled reward and value predictions to handle scale variation, layer normalization~\citep{ba2016layer} for improved robustness, and return normalization using an exponential moving average (EMA) of return percentiles. With these changes, DreamerV3 achieved state-of-the-art performance across a wide range of benchmarks, often outperforming both specialized model-free and other model-based methods.

In parallel to the Dreamer line of work, other approaches have explored alternative world model designs. Beyond recurrent architectures, Transformer-based world models have been proposed to improve long-term memory and representation learning~\citep{micheli2022transformers, chen2021transdreamer, robine2023transformer, zhang2023storm}. Diffusion-based world models~\citep{sohl2015deep} have also emerged as a compelling alternative to discrete latent variable sequence modeling, demonstrating strong performance across a range of settings~\citep{alonso2024diffusion, valevski2025diffusion, bar2025navigation, hafner2025training, hu2026multiplayer}. Another line of work has investigated multitask world models, training a single world model jointly across multiple tasks to improve generalization and achieve stronger performance than single-task models~\citep{hansen2024td, hansen2026learning}. Model-based reinforcement learning approaches using Monte-Carlo tree search (MCTS)~\citep{coulom2006efficient} have also been explored following the success of Alpha Go~\citep{silver2017mastering, silver2018general, schrittwieser2020mastering, ye2021mastering, wang2024efficientzero}.

\subsubsection{Memory in World Models}

Multiple approaches have focused on improving the memory capabilities of world models. \citet{pasukonis2023evaluating} proposed to augment Dreamer with Truncated Backpropagation Through Time (TBTT), conditioning training sequences on previous context representations to preserve information over time. TransDreamer~\citep{chen2021transdreamer} replaced Dreamer’s RSSM by a Transformer State-Space Model (TSSM) using masked self-attention to process past context more easily. The agent was evaluated on Hidden Order Discovery tasks requiring long-term memory and reasoning, demonstrating better performance. Recall To Imagine (R2I)~\citep{samsami2024mastering} proposed the Structured State-Space Model (S3M) to improve computational efficiency when handling long sequences. This allows R2I to process very long sequences and achieve strong performance on the Memory Maze benchmark. Concurrently,~\citet{deng2023facing} proposed S4WM, demonstrating that the S4 architecture outperforms Transformer-based
world models in terms of long-term memory. Alternatively, EDELINE~\citep{lee2025edeline} proposed to augment the DIAMOND world model~\citep{alonso2024diffusion} with a Mamba~\citep{gu2023mamba} SSM to enhance memory capabilities. A similar approach was concurrently proposed in~\citet{savov2025statespacediffuser} where the authors use Mamba to improve model predictions on CSGO~\citep{pearce2022counter}.

\subsubsection{Test-Time Training}

Test-Time Training (TTT) was originally proposed in~\citet{sun2020test} for improving the performance of predictive models when training and test data come from different distributions. This work proposed to turn a single unlabeled test sample into a self-supervised learning problem, on which to update the model parameters before making a prediction. The initial TTT paper used image rotation prediction~\citep{gidaris2018unsupervised} as a self-supervised task to update model weights and achieve higher classification accuracy on corrupted test images. The method was later improved with TTT-MAE~\citep{gandelsman2022test}, using MAE~\citep{he2022masked} instead of rotation prediction to reach higher performance. Test-time training was subsequently applied to multiple domains like reinforcement learning~\citep{hansen2021self}, language modeling~\citep{hardt2023test}, and video processing~\citep{wang2025test}. In the area of reinforcement learning, \citet{hansen2021self} proposed Policy Adaptation during Deployment (PAD)~\citep{hansen2021self}, applying TTT to model-free reinforcement learning algorithms~\citep{haarnoja2018soft, mnih2016asynchronous}. They demonstrated that PAD improves performance under distribution shifts on the DeepMind Control~\citep{tassa2018deepmind} and CRLMaze~\citep{lomonaco2020continual} benchmarks. They also showed that PAD enables more effective sim-to-real transfer for robotic manipulation tasks.

%% file: tables/study_objective.tex
\begin{table}[ht]
    \caption{Ablation study on Test-Time Optimization objective.}
    \centering
    \setlength{\tabcolsep}{4.5pt}
    \hfill \break
    \begin{tabular}{lccccc}
    \toprule
    \textbf{Method} & \textbf{PAD} & \textbf{TT-MBRL} & \textbf{w/o KL loss} & \textbf{w/o rwd-cont loss} & \textbf{w/o actor-critic loss} \\
    \midrule
    DreamerV3 & $\hphantom{0}25.5$ & $185.0$ & $\hphantom{0}12.3$ & $\mathbf{187.9}$ & $\hphantom{0}47.8$ \\
    TWISTER & $\hphantom{0}35.5$ & $240.1$ & $\hphantom{0}12.9$ & $\mathbf{240.9}$ & $\hphantom{0}50.2$  \\
    R2I & $\hphantom{0}42.3$ & $\mathbf{244.3}$ & $\hphantom{0}14.9$ & $242.7$ & $158.5$ \\
    \bottomrule
    \end{tabular}
    \label{table:study_objective}
\end{table}

%% file: tables/study_train_test_ratio.tex
\begin{table}[ht]
    \caption{Ablation study on TT-MBRL train/test sampling ratio.}
    \centering
    \setlength{\tabcolsep}{4.5pt}
    \hfill \break
    \begin{tabular}{lcccccc}
    \toprule
    \textbf{Method} & \textbf{w/o TT-MBRL} & \textbf{100/0 (train-only)} & \textbf{75/25} & \textbf{50/50} & \textbf{25/75} & \textbf{0/100 (test-only)} \\
    \midrule
    DreamerV3 & $\hphantom{0}46.2$ & $\hphantom{0}47.0$ & $160.9$ & $\mathbf{185.0}$ & $179.4$ & $117.6$ \\
    TWISTER & $\hphantom{0}72.0$ & $\hphantom{0}69.2$ & $\mathbf{244.5}$ & $240.1$ & $235.2$ & $167.6$ \\
    R2I & $171.6$ & $171.7$ & $242.5$ & $244.3$ & $\mathbf{249.7}$ & $218.8$ \\
    \bottomrule
    \end{tabular}
    \label{table:study_train_test_ratio}
\end{table}

%% file: tables/study_frequency.tex
\begin{table}[ht]
    \caption{Study on test-time gradient update frequency.}
    \centering
    \setlength{\tabcolsep}{4pt}
    \hfill \break
    \begin{tabular}{lcccccccc}
    \toprule
    \textbf{Policy Step/Grad Step} & \textbf{w/o TT-MBRL} & \textbf{32} & \textbf{16} & \textbf{8} & \textbf{4} & \textbf{2} & \textbf{1}  & \textbf{1/2} \\
    \midrule
    DreamerV3 & $\hphantom{0}46.2$ & $\hphantom{0}79.9$ & $107.3$ & $132.1$ & $152.9$ & $165.7$ & $\mathbf{185.0}$ & $178.7$ \\
    TWISTER & $\hphantom{0}72.0$ & $108.9$ & $139.8$ & $178.3$ & $198.5$ & $225.8$ & $240.1$ & $\mathbf{245.6}$ \\
    R2I & $171.6$ & $182.3$ & $190.2$ & $205.8$ & $213.6$ & $230.9$ & $244.3$ & $\mathbf{251.8}$ \\
    \bottomrule
    \end{tabular}
    \label{table:study_frequency}
\end{table}